\documentclass[journal]{IEEEtran}

\ifCLASSINFOpdf

\else

\fi

\usepackage[colorlinks, linkcolor=black, urlcolor=blue,anchorcolor=black, citecolor=blue]{hyperref}
\usepackage{amsmath, amssymb}
\usepackage{algorithm}
\usepackage{algorithmic}
\usepackage{booktabs}
\usepackage{dsfont}
\usepackage[dvips]{epsfig}
\usepackage{epsfig}
\usepackage{float}
\usepackage{graphicx}
\usepackage{lettrine}
\usepackage[mathlines]{lineno}
\usepackage{multirow}
\usepackage{times}
\usepackage[tight,footnotesize]{subfigure}
\usepackage[mathlines]{lineno}
\usepackage{xcolor}
\usepackage{url}

\newtheorem{theorem}{\bf Theorem}
\newtheorem{definition}{\bf Definition}
\newtheorem{lemma}{\bf Lemma}

\begin{document}
\title{Constructing the L2-Graph for Robust Subspace Learning and Subspace Clustering}

\author{Xi Peng,
        Zhiding Yu,
        Huajin Tang,~\IEEEmembership{Member,~IEEE,}
        and Zhang Yi,~\IEEEmembership{Senior Member,~IEEE}
\IEEEcompsocitemizethanks{\IEEEcompsocthanksitem Xi Peng and Huajin Tang are with Institute for Infocomm Research, Agency for Science, Technology and Research (A*STAR), Singapore 138632; Zhiding Yu is with Department of Electrical and Computer Engineering, Carnegie Mellon University, Pittsburgh, PA, 15213, USA;
Zhang Yi is with the Machine Intelligence Laboratory, College of Computer
Science, Sichuan University, Chengdu 610065, P.R. China. (E-mail: pangsaai@gmail.com; yzhiding@andrew.cmu.edu; htang@i2r.a-star.edu.sg; zhangyi@scu.edu.cn).}
\thanks{Corresponding author: Huajin Tang (htang@i2r.a-star.edu.sg)}}

\markboth{}%
{}

\IEEEcompsoctitleabstractindextext{\begin{abstract}
Under the framework of graph-based learning, the key to robust subspace clustering and subspace learning is to obtain a good similarity graph that eliminates the effects of errors and retains only connections between the data points from the same subspace (i.e., intra-subspace data points).  Recent works achieve good performance by modeling errors into their objective functions to remove the errors from the inputs. However, these approaches face the limitations that the structure of errors should be known prior and a complex convex problem must be solved. In this paper, we present a novel method to eliminate the effects of the errors from the projection space (representation) rather than from the input space. We first prove that $\ell_1$-, $\ell_2$-, $\ell_{\infty}$-, and nuclear-norm based linear projection spaces share the property of Intra-subspace Projection Dominance (IPD), i.e., the coefficients over intra-subspace data points are larger than those over inter-subspace data points. Based on this property, we introduce a method to construct a sparse similarity graph, called L2-Graph. The subspace clustering and subspace learning algorithms are developed upon L2-Graph. Experiments show that L2-Graph algorithms outperform the state-of-the-art methods for feature extraction, image clustering, and motion segmentation in terms of accuracy, robustness, and time efficiency.
\end{abstract}

}

\maketitle

\IEEEdisplaynotcompsoctitleabstractindextext
\IEEEpeerreviewmaketitle


\section{Introduction}
\label{sec:1}

\IEEEPARstart{T}{he} key to graph-based learning algorithms is the sparse eigenvalue problem, i.e., constructing a block-diagonal affinity matrix whose nonzero entries correspond to the data points belonging to the same subspace (i.e., intra-subspace data points). Based on the affinity matrix, a series of algorithms~\cite{Roweis2000,Belkin2003,Yan2007} can be derived for various tasks such as subspace clustering and subspace learning.

Currently, there are two popular ways to build a similarity graph, one is based on pairwise distances (e.g., Euclidean distance) and the other is based on reconstruction coefficients (e.g., sparse representation). The second family of methods has recently attracted a lot of interest from the community, where one assumes that each data point can be represented as a linear combination of other points. When the data is clean and the subspaces are mutually independent or disjoint, the approaches such as~\cite{Costeira1998, He2005Lap} are able to
well handle the subspace clustering and subspace learning problems. In real applications, however, the data sets are likely to contain various types of noise and data could often lie near the intersection of multiple dependent subspaces. As a result, inter-subspace data points (i.e., the data points with different labels) may connect to each other with very high edge weights, which degrades the performance of graph-based methods. To achieve more robust results, some algorithms have been proposed~\cite{Vidal2005,Yan2006,Chen2009,Rao2010,Yan2009}. In~\cite{Vidal2011}, Vidal conducted a comprehensive survey regarding subspace clustering.

Recently,~\cite{Elhamifar2013,Qiao2010,Cheng2010,Liu2013}
provided a new way to construct the graph using the sparsest or lowest-rank representation. Moreover,~\cite{Elhamifar2013,Liu2013} remove errors from the inputs by modeling the errors in their objective functions. Both theoretical analysis and experimental results show that the methods can handle certain specific types of errors and have achieved good performance. Inspired by their success, the error-removing method is widely adopted in a number of approaches~\cite{Liu2012b,Liu2011,Favaro2011,peng2013scalable,Wang2013,Chen2014SMC,Peng2014SMC}.

One major limitation of these approaches is that the structure of errors should be known as the prior knowledge so that the errors can be formulated into the objective function. In practice, this prior knowledge is difficult to get and the algorithms may work well only if the adopted assumption is consistent with the true structure of the errors. Moreover, these methods must solve a convex problem whose computational complexity is at least proportional to the cubic of the data size.

Different from these approaches, we propose a novel error-removing method which aims to eliminate the effect of errors from the $\ell_p$-norm- and nuclear-norm-based projection space (i.e., encoding and then removing errors), where $p=\{1,2,\infty\}$. The method is based on a mathematically trackable property of the projection space, i.e., Intra-subspace Projection Dominance (IPD). Based on our theoretical result, we further propose L2-Graph for subspace clustering and subspace learning by considering the case of $\ell_2$-norm. The proposed method can handle various errors even though the structure of errors is unknown and the data are grossly corrupted.

The contributions of this paper is summarized as follows: 1) We prove the property of IPD shared by $\ell_{1}$-, $\ell_{2}$-, $\ell_{\infty}$-, and nuclear-norm-based projection space, i.e., the coefficients with small values (trivial coefficients) always correspond to the projections over the errors. 2) We propose a graph-building method based on $\ell_2$-norm, named L2-Graph. The method has a closed-form solution and is more efficient than most existing methods such as~\cite{Elhamifar2013,Qiao2010,Cheng2010,Liu2013,Liu2011}.
 3) We incorporate L2-Graph into the graph embedding framework~\cite{Yan2007,Ng2002,He2005} and develop two  new algorithms for robust subspace clustering and subspace learning.

The paper is an extension of the work in~\cite{peng2015robust}. Compared with~\cite{peng2015robust}, we further improve our work from the following several aspects: 1) Besides $\ell_1$-, $\ell_2$-, and $\ell_{\infty}$-norm based projection space, we prove that nuclear-norm-based projection space also possesses the property of IPD; 2) Motivated by the success of sparse representation in subspace learning~\cite{Yan2009,Qiao2010,Cheng2010,Lu2015}, we propose a new subspace learning method derived upon the L2-Graph. Extensive experimental results show that our method outperform state-of-the-art feature extraction method in accuracy and robustness; 3) We explore the potential of L2-Graph in estimating the latent structures of data space. 4) Besides image clustering, we extend L2-Graph in the applications of motion segmentation and unsupervised feature extraction; 5) We investigate the performance of our method more thoroughly (8 new data sets); 6) We conduct  compressive analysis for our method, including the effect of different parameters, different errors (additive and non-additive noises and partial disguises), and different experimental settings.

The rest of the article is organized as follows: Section \ref{sec:2} presents some related works on graph construction methods. Section~\ref{sec:3} prove that it is feasible to eliminate the effects of errors from the representation. Section~\ref{sec:3} proposes the L2-Graph algorithm and  two methods for subspace learning and subspace clustering derived upon L2-Graph. Section~\ref{sec:5} reports the performance of the proposed methods in the context of feature extraction, image clustering, and motion segmentation. Finally, Section~\ref{sec:6} summarizes this work.

\textbf{Notations:} Unless specified otherwise, \textbf{lower-case bold letters} represent column vectors and \textbf{upper-case bold ones} represent matrices. $\mathbf{A}^T$ and $\mathbf{A}^{-1}$ denote the transpose and pseudo-inverse of the matrix $\mathbf{A}$, respectively. $\mathbf{I}$ denotes the identity matrix. \tablename~\ref{tab1} summarizes some notations and abbreviations used throughout the paper.

\begin{table}[t]
\caption{Notations and Abbreviations.}
\label{tab1}
\begin{center}
\begin{small}
\begin{tabular}{ll}
\toprule
Notation (Abbr.) & Definition\\
\midrule
$n$ & data size\\
$m$ & the dimension of samples\\
$m^{\prime}$ & the dimension of features\\
$r$ & the rank of a given matrix\\
$c$ & the number of subspace\\
$k$ & the neighborhood size\\
$\mathbf{x}\in \mathds{R}^{m}$ & a data point\\
$\mathbf{c}\in \mathds{R}^{n}$ & the representation of $\mathbf{x}$ over $\mathbf{D}$\\
$\mathbf{D}=[\mathbf{d}_1, \mathbf{d}_2, \ldots, \mathbf{d}_n]$ & a given dictionary\\
$\mathbf{D}_{x}\in \mathbf{D}$ & $\mathbf{x}$ and $\mathbf{D}_{x}$ have the same labels\\
$\mathbf{D}_{-x}$ & the data points of $\mathbf{D}$ except $\mathbf{D}_{x}$\\
$\mathbf{D}=\mathbf{U} \mathbf{\Sigma} \mathbf{V}^{T}$ & full  SVD of $\mathbf{D}$\\
$\mathbf{D}=\mathbf{U}_{r} \mathbf{\Sigma}_{r} \mathbf{V}^{T}_{r}$ & skinny SVD of  $\mathbf{D}$\\
IPD & Intra-subspace projection dominance\\
LLR & Locally Linear Representation\\
SR & Sparse Representation \\
LRR & Low rank representation\\
SVD & Singular value decomposition\\
\bottomrule
\end{tabular}
\end{small}
\end{center}
\end{table}

\section{Related Work}
\label{sec:2}

Over the past two decades, a number of graph-based algorithms have been proposed with various applications such as feature extraction~\cite{He2005}, subspace clustering~\cite{Luxburg2007}, and object tracking~\cite{Papadakis2011}. The key to these algorithms is the construction of the similarity graph and the performance of the algorithms largely hinges on whether the graph can accurately determine the neighborhood of each data point, particularly when the data set contains errors.

There are two ways to build a similarity graph, i.e., the pairwise distance and the reconstruction coefficients. In the pairwise distance setting, one of the most popular metric is Euclidean distance with Heat Kernel, i.e.,
\begin{equation}
\label{sec2:equ1}
\mathrm{similarity}(\mathbf{x}_{i},\mathbf{x}_{j})=\exp^{-\frac{\|\mathbf{x}_{i}-\mathbf{x}_{j}\|_{2}^{2}}{\tau}},
\end{equation}
where $\mathbf{x}_{i}$ and $\mathbf{x}_{j}$ denote two data points and $\tau$ denotes the width of the Heat Kernel.

This metric has been used to build the similarity graph for subspace clustering~\cite{Ng2002} and subspace learning~\cite{He2005Lap}. However, pairwise distance is sensitive to noise  and outliers since its value only depends on the corresponding two data points. Consequently, pairwise distance based algorithms may fail to handle noise corrupted data.

Alternatively, reconstruction coefficients  based similarity is data-adaptive. Such property benefits the robustness, and as a result these algorithms have become increasingly popular, especially in high-dimensional data analysis. Three reconstruction coefficients are widely used to represent the neighbor relations among data points, i.e., Locally Linear Representation (LLR)~\cite{Roweis2000}, Sparse Representation (SR), and Low Rank Representation (LRR).

For each data point $\mathbf{x}_i$, LLR seeks to solve the following optimization problem
\begin{equation}
\label{sec2:equ2}
\min\hspace{1mm}\|\mathbf{x}_i-\mathbf{D}_i \mathbf{c}_{i}\|_2^2
\hspace{3mm} \mathrm{s.t.} \hspace{1mm}\mathbf{1}^T\mathbf{c}_{i}=1,
\end{equation}
where $\mathbf{c}_i\in\mathds{R}^{k}$ is the coefficient of $\mathbf{x}_i$ over $\mathbf{D}_i\in\mathds{R}^{m\times k}$ and $\mathbf{D}_i$ consists of $k$ nearest neighbors of $\mathbf{x}_i$ in terms of Euclidean distance. Another well known relevant work is Neighborhood Preserving Embedding (NPE)~\cite{He2005} which uses LLR to construct the similarity graph for subspace learning. A significant problem associated with such methods is that they cannot achieve a good result unless the data are uniformly sampled from a smooth manifold. Moreover, if the data are grossly corrupted, the performance of these methods will degrade considerably.

Different from LLR, SR uses a few bases to represent each data point. Such strategy is widely used to construct the similarity graph for subspace clustering~\cite{Elhamifar2013,Cheng2010} and subspace learning~\cite{Qiao2010,Cheng2010}. A robust version of SR is
\begin{align}
\label{sec2:equ3}
&\underset{\mathbf{C},\mathbf{E},\mathbf{Z}}{\min} \hspace{1mm} \|\mathbf{C}\|_1+\lambda_{E}\|\mathbf{E}\|_1+\lambda_{Z}\|\mathbf{Z}\|_F \notag\\
&\mathrm{s.t.} \hspace{1mm} \mathbf{X}=\mathbf{X} \mathbf{C}+\mathbf{E}+\mathbf{Z}, \mathbf{1}^{T}\mathbf{C}=\mathbf{1}, \mathrm{diag}(\mathbf{C})=0,
\end{align}
where $\mathbf{X}\in \mathds{R}^{m\times n}$ is the given data set, $\mathbf{C}\in \mathds{R}^{n\times n}$ denotes the sparse representation of the data set $\mathbf{X}$, $\mathbf{E}$ corresponds to the sparse outlying entries and $\mathbf{Z}$ denotes the reconstruction errors caused by the constrained representation flexibility. $\mathbf{1}$ is a column vector with $n$ entries of $1$, and the parameters $\lambda_{E}$ and $\lambda_{Z}$ balance the cost terms of the objective function.

Different from SR, LRR uses the low rank representation to build the graph, which is proved to be very effective in subspace clustering~\cite{Liu2013} and subspace learning~\cite{Liu2011}. The method solves the following optimization problem:
\begin{equation}
\label{sec2:equ4}
\min\hspace{1mm}\|\mathbf{C}\|_{\ast}+\lambda\|\mathbf{E}\|_{p}
\hspace{3mm} \mathrm{s.t.} \hspace{1mm} \mathbf{X}=\mathbf{XC}+\mathbf{E},
\end{equation}
where $\|\cdot\|_{\ast}$ denotes the nuclear norm that summarizes the singular value of a given data matrix. $\|\cdot\|_{p}$ could be chosen as $\ell_{2,1}$-, $\ell_{1}$-, or Frobenius-norm. The choice of the norm only depends on which kind of error is assumed in the data set. Specifically, $\ell_{2,1}$-norm is usually adopted to depict sample-specific corruption and outliers, $\ell_1$-norm is used to characterize random corruption, and Frobenius norm is used to describe the Gaussian noise.

From (\ref{sec2:equ3}) and (\ref{sec2:equ4}), it is easy to see that SR and LRR based methods remove errors from the input space by modeling them in their objective functions. A number of works~\cite{Liu2012b,Liu2011,Favaro2011,Wang2013}
 have also adopted such error-removing strategy, showing its effectiveness in various applications. In this paper, we propose a novel error-removing method that seeks to eliminate the effect of errors from the projection space instead of the input space. The method is mathematically trackable and does not suffer from the limitation of error structure estimation as most existing methods do.

\section{Intra-subspace Projection Dominance}
\label{sec:3}

In this section, we prove that the coefficients over intra-subspace data points is larger than those over inter-subspace data points in $\ell_p$- and nuclear-norm based projection space, namely, Intra-subspace Projection Dominance. Data points are called \textit{intra-subspace data points} if they are from the same subspace, and otherwise \textit{inter-subspace data points}.

\subsection{IPD in $\ell_p$-norm based Projection Space}
\label{sec3.1}

Let $\mathbf{x}\ne \mathbf{0}$ be a data point drawn from the union of subspaces (denoted by $\mathcal{S}_{\mathbf{D}}$) that is spanned by $\mathbf{D}=[\mathbf{D}_x\ \mathbf{D}_{-x}]$, where $\mathbf{D}_{x}$ and $\mathbf{D}_{-x}$ consist of the intra-cluster and inter-cluster data points of $\mathbf{x}$, respectively. Note that under our setting, noise and outliers are regarded as inter-cluster data points of $\mathbf{x}$. Without loss of generality, let
$\mathcal{S}_{\mathbf{D}_{x}}$ and $\mathcal{S}_{\mathbf{D}_{-x}}$ be the subspace spanned by
$\mathbf{D}_{x}$ and $\mathbf{D}_{-x}$, respectively. Hence, there are only two possibilities for the location of $\mathbf{x}$, i.e., in the intersection between $\mathcal{S}_{\mathbf{D}_{x}}$ and $\mathcal{S}_{\mathbf{D}_{-x}}$ (denoted as $\mathbf{x}\in
\{\mathcal{S}|\mathcal{S}=\mathcal{S}_{\mathbf{D}_{x}}\cap
\mathcal{S}_{\mathbf{D}_{-x}}\}$), or in $\mathcal{S}_{\mathbf{D}_{x}}$
except the intersection (denoted as $\mathbf{x}\in
\{\mathcal{S}|\mathcal{S}=\mathcal{S}_{\mathbf{D}_{x}}\backslash
\mathcal{S}_{\mathbf{D}_{-x}}\}$).

Let $\mathbf{c}_{\mathbf{D}_{x}}^\ast$ and $\mathbf{c}_{\mathbf{D}_{-x}}^\ast$ be the optimal solutions of
\begin{equation}
\label{theo:equ1}
\min\hspace{1mm}\|\mathbf{c}\|_p \hspace{3mm}
\mathrm{s.t.}\hspace{1mm} \mathbf{x}=\mathbf{D}\mathbf{c},
\end{equation}
over $\mathbf{D}_{x}$ and $\mathbf{D}_{-x}$, respectively. $\|\cdot\|_p$ denotes the $\ell_p$-norm, $p=\{1,2,\infty\}$. We aim to investigate the conditions under which, for every nonzero data point $\mathbf{x}\in \mathcal{S}_{\mathbf{D}_{x}}$ satisfying $\|\mathbf{c}_{\mathbf{D}_{x}}^{\ast}\|_{p} < \|\mathbf{c}_{\mathbf{D}_{-x}}^{\ast}\|_{p}$, the coefficients over intra-subspace data points are larger than those over inter-subspace data points, i.e., $[\mathbf{c}_{\mathbf{D}_{x}}^\ast]_{r_{x},1}>[\mathbf{c}_{\mathbf{D}_{-x}}^\ast]_{1,1}$ (IPD property).  Here, $[\mathbf{c}_{\mathbf{D}_{x}}^\ast]_{r_{x},1}$ denotes the $r_{x}$-th largest absolute value of the entries of $\mathbf{c}_{\mathbf{D}_{x}}^\ast$, and $r_{x}$ is the dimensionality of $\mathcal{S}_{\mathbf{D}}$.

In the following analysis, Theorem~\ref{lem1} and Theorem~\ref{lem3} show $[\mathbf{c}_{\mathbf{D}_{x}}^\ast]_{r_{x},1}>[\mathbf{c}_{\mathbf{D}_{-x}}^\ast]_{1,1}$ when $\mathbf{x}\in
\{\mathcal{S}|\mathcal{S}=\mathcal{S}_{\mathbf{D}_{x}}\backslash
\mathcal{S}_{\mathbf{D}_{-x}}\}$ and $\mathbf{x}\in
\{\mathcal{S}|\mathcal{S}=\mathcal{S}_{\mathbf{D}_{x}}\cap
\mathcal{S}_{\mathbf{D}_{-x}}\}$, respectively. Lemma~\ref{lem2} and Definition~\ref{def1} are preliminary steps toward Theorem~\ref{lem3}.

\begin{theorem}
\label{lem1}
For any nonzero data point $\mathbf{x}$ in the
subspace $\mathcal{S}_{\mathbf{D}_{x}}$ except the intersection
between $\mathcal{S}_{\mathbf{D}_{x}}$ and
$\mathcal{S}_{\mathbf{D}_{-x}}$, i.e., $\mathbf{x}\in
\{\mathcal{S}|\mathcal{S}=\mathcal{S}_{\mathbf{D}_{x}}\backslash
\mathcal{S}_{\mathbf{D}_{-x}}\}$, we must have
$[\mathbf{c}_{\mathbf{D}_{x}}^{\ast}]_{r_{x},1}>[\mathbf{c}_{\mathbf{D}_{-x}}^{\ast}]_{1,1}$, where
$\mathbf{c}^{\ast}=\begin{bmatrix}\mathbf{c}_{\mathbf{D}_{x}}^{\ast}\\\mathbf{c}_{\mathbf{D}_{-x}}^{\ast}
\end{bmatrix}$ is the optimal solution of (\ref{theo:equ1}) and is partitioned according to the data set $\mathbf{D} = [\mathbf{D}_{x}\ \mathbf{D}_{-x}]$.
\end{theorem}

\begin{proof}
For the nonzero data point $\mathbf{x}$, suppose there exists a nonzero vector $\mathbf{c}_{\mathbf{D}_{-x}}^{\ast}$ such that
\begin{equation}
\mathbf{x} = \mathbf{D}_{x}\mathbf{c}_{\mathbf{D}_{x}}^{\ast}+\mathbf{D}_{-x}\mathbf{c}_{\mathbf{D}_{-x}}^{\ast},
\end{equation}
then
\begin{equation}
\mathbf{x} - \mathbf{D}_{x}\mathbf{c}_{\mathbf{D}_{x}}^{\ast}=\mathbf{D}_{-x}\mathbf{c}_{\mathbf{D}_{-x}}^{\ast},
\end{equation}

Since $\mathbf{x}\in\mathcal{S}_{\mathbf{D}_{x}}$, then $\mathbf{x} - \mathbf{D}_{x}\mathbf{c}_{\mathbf{D}_{x}}^{\ast}\in \mathcal{S}_{\mathbf{D}_{x}}$, i.e., $\mathbf{D}_{-x}\mathbf{c}_{\mathbf{D}_{-x}}^{\ast}\in \mathcal{S}_{\mathbf{D}_{x}}$.

As $\mathcal{S}_{\mathbf{D}_{x}} \cap \mathcal{S}_{\mathbf{D}_{-x}}=0$ and $\mathbf{x}\ne \mathbf{0}$, then $\mathbf{c}_{\mathbf{D}_{-x}}^\ast=\mathbf{0}$ and $\mathbf{c}_{\mathbf{D}_{x}}^\ast \ne \mathbf{0}$. This contradicts the assumption $\mathbf{c}_{\mathbf{D}_{-x}}^{\ast}\ne \mathbf{0}$. Then, we must have $\mathbf{c}_{\mathbf{D}_{x}}^{\ast}\ne \mathbf{0}$ and $\mathbf{c}_{\mathbf{D}_{-x}}^{\ast}=\mathbf{0}$ which implies that $[\mathbf{c}_{\mathbf{D}_{x}}^{\ast}]_{r_{0},1}>[\mathbf{c}_{\mathbf{D}_{-x}}^{\ast}]_{1,1}$.

This completes the proof.
\end{proof}

\begin{lemma}
\label{lem2}
Consider a  nonzero data point $\mathbf{x}\in \mathcal{S}_{\mathbf{D}_{x}}$ and $\mathbf{x}$ lies in the intersection between $\mathcal{S}_{\mathbf{D}_{x}}$ and $\mathcal{S}_{\mathbf{D}_{-x}}$, i.e., $\mathbf{x}\in \{\mathcal{S}|\mathcal{S}=\mathcal{S}_{\mathbf{D}_{x}}\cap \mathcal{S}_{\mathbf{D}_{-x}}\}$. Let $\mathbf{c}^{\ast}$, $\mathbf{z}_{\mathbf{D}_{x}}$, and $\mathbf{z}_{\mathbf{D}_{-x}}$ be the optimal solution of
\begin{equation}
\label{lem2equ1}
\min\hspace{1mm}\|\mathbf{c}\|_p \hspace{3mm} \mathrm{s.t.}\hspace{1mm} \mathbf{x}=\mathbf{D}\mathbf{c}
\end{equation}
over $\mathbf{D}$, $\mathbf{D}_{x}$, and $\mathbf{D}_{-x}$, and $\mathbf{c}^{\ast}=\begin{bmatrix}\mathbf{c}_{\mathbf{D}_{x}}^{\ast}\\\mathbf{c}_{\mathbf{D}_{-x}}^{\ast} \end{bmatrix}$ is partitioned according to the sets $\mathbf{D}=[\mathbf{D}_{x}\ \mathbf{D}_{-x}]$. If  $\|\mathbf{z}_{\mathbf{D}_{x}}\|_p < \|\mathbf{z}_{\mathbf{D}_{-x}}\|_p$, then $\mathbf{c}_{\mathbf{D}_{-x}}^{\ast}=\mathbf{0}$ so that $[\mathbf{c}_{\mathbf{D}_{x}}^{\ast}]_{r_{x},1}>[\mathbf{c}_{\mathbf{D}_{-x}}^{\ast}]_{1,1}$.
\end{lemma}

\begin{proof}
($\Longleftarrow$) We prove the result using contradiction. Assume $\mathbf{c}_{\mathbf{D}_{-x}}^{\ast}\ne\mathbf{0}$, then
\begin{equation}
\label{lem2equ2}
\mathbf{x}-\mathbf{D}_{x}\mathbf{c}_{\mathbf{D}_{x}}^{\ast}=\mathbf{D}_{-x}\mathbf{c}_{\mathbf{D}_{-x}}^{\ast}.
\end{equation}

Define $\mathbf{y}=\mathbf{x}-\mathbf{D}_{x}\mathbf{c}_{\mathbf{D}_{x}}^{\ast}$. Since $\mathbf{x}\in\mathcal{S}_{\mathbf{D}_{x}}$, then $\mathbf{y}$ must belong to $\mathcal{S}_{\mathbf{D}_{x}}$. Thus,

\begin{equation}
\label{lem2equ3}
\mathbf{x}=\mathbf{D}_{x}\mathbf{c}_{\mathbf{D}_{x}}^{\ast}+\mathbf{D}_{x}\mathbf{z}_{\mathbf{D}_{x}}.
\end{equation}

Moreover, the right side of (\ref{lem2equ2}) corresponds to the data point that lies in $\mathcal{S}_{\mathbf{D}_{-x}}$, then we have

\begin{equation}
\label{lem2equ4}
\mathbf{x}=\mathbf{D}_{x}\mathbf{c}_{\mathbf{D}_{x}}^{\ast}+\mathbf{D}_{-x}\mathbf{z}_{\mathbf{D}_{-x}},
\end{equation}

Clearly, $\begin{bmatrix} \mathbf{c}_{\mathbf{D}_{x}}^{\ast}+\mathbf{z}_{\mathbf{D}_{x}} \\ \mathbf{0}\end{bmatrix}$ and $\begin{bmatrix} \mathbf{c}_{\mathbf{D}_{x}}^{\ast} \\ \mathbf{z}_{\mathbf{D}_{-x}}\end{bmatrix}$ are feasible solutions of (\ref{lem2equ1}) over $[\mathbf{D}_{x}\ \mathbf{D}_{-x}]$. According to the triangle inequality and the condition $\|\mathbf{z}_{\mathbf{D}_{x}}\|_p < \|\mathbf{z}_{\mathbf{D}_{-x}}\|_p$, we have
\begin{equation}
\label{lem2equ5}
\left\| \begin{bmatrix} \mathbf{c}_{\mathbf{D}_{x}}^{\ast}+\mathbf{z}_{\mathbf{D}_{x}} \\ \mathbf{0}\end{bmatrix} \right\|_p
\leq \|\mathbf{c}_{\mathbf{D}_{x}}^{\ast}\|_p+\|\mathbf{z}_{\mathbf{D}_{x}}\|_p
<    \|\mathbf{c}_{\mathbf{D}_{x}}^{\ast}\|_p+\|\mathbf{z}_{\mathbf{D}_{-x}}\|_p.
\end{equation}

From (\ref{lem2equ4}), we have $\|\mathbf{z}_{\mathbf{D}_{-x}}\|_p \leq \|\mathbf{c}_{\mathbf{D}_{-x}}^{\ast}\|_p$ as $\|\mathbf{z}_{\mathbf{D}_{-x}}\|_p$ is the optimal solution of (\ref{lem2equ1}) over $\mathbf{D}_{-x}$. Then, $\left\| \begin{bmatrix} \mathbf{c}_{\mathbf{D}_{x}}^{\ast}+\mathbf{z}_{\mathbf{D}_{x}} \\ \mathbf{0}\end{bmatrix} \right\|_p < \left\|\begin{bmatrix}\mathbf{c}_{\mathbf{D}_{x}}^{\ast}\\\mathbf{c}_{\mathbf{D}_{-x}}^{\ast}\end{bmatrix} \right\|_p$. It contradicts the fact that $\left\|\begin{bmatrix}\mathbf{c}_{\mathbf{D}_{x}}^{\ast}\\\mathbf{c}_{\mathbf{D}_{-x}}^{\ast}\end{bmatrix} \right\|_p$ is the optimal solution of (\ref{lem2equ1}) over $\mathbf{D}$.

($\Longrightarrow$) We prove the result using contradiction. For a nonzero data point $\mathbf{x}\in \{\mathcal{S}|\mathcal{S}=\mathcal{S}_{\mathbf{D}_{x}}\cap \mathcal{S}_{\mathbf{D}_{-x}}\}$, assume $\|\mathbf{z}_{\mathbf{D}_{x}}\|_p \geq \|\mathbf{z}_{\mathbf{D}_{-x}}\|_p$. Thus, for the data point $\mathbf{y}=\mathbf{x}$, (\ref{lem2equ1}) will only choose the points from $\mathcal{S}_{\mathbf{D}_{-x}}$ to represent $\mathbf{x}$. This contradicts to $\mathbf{c}_{\mathbf{D}_{x}}^{\ast}\ne \mathbf{0}$ and $\mathbf{c}_{\mathbf{D}_{-x}}^{\ast}=\mathbf{0}$.

This completes the proof.
\end{proof}

Lemma~\ref{lem2} provides the necessary and sufficient condition to guarantee the IPD property of $\ell_{p}$-norm based projection space, but it does not bridge the relationship between IPD and the data distribution.
To establish such relationship, we measure the distance among the subspaces $\mathcal{S}_{\mathbf{D}_{x}}$ and $\mathcal{S}_{\mathbf{D}_{-x}}$ using the first principle angle $\theta_{\min}$ and show the IPD property under such setting. Moreover, we derive a more relaxed condition which is more easily satisfied in practice.

\begin{definition}[The First Principal Angle]
\label{def1}
Let $\xi$ be a Euclidean vector-space, and consider the two
subspaces $\mathcal{W}$, $\mathcal{V}$ with $\mathrm{dim}(\mathcal{W}):=r_{\mathcal{W}}\leq \mathcal{V}:=r_{\mathcal{V}}$. There exists a set of
angles $\{\theta_i\}_{i=1}^{r_{\mathcal{W}}}$ called the principal angles, the first one being defined as:
\begin{equation}
\label{theo:equ2}
\theta_{\min}:=\min_{\mathbf{\mu},\mathbf{\nu}}\left\{\arccos\left(\frac{\mathbf{\mu}^{T}\mathbf{\nu}}{\|\mathbf{\mu}\|_2\|\mathbf{\nu}\|_2}\right)\right\},
\end{equation}
where $\mathbf{\mu}\in \mathcal{W}$ and $\mathbf{\nu}\in \mathcal{V}$.
\end{definition}

\begin{theorem}
\label{lem3}
Consider the nonzero data point $\mathbf{x}$ in the intersection between $\mathcal{S}_{\mathbf{D}_{x}}$ and $\mathcal{S}_{\mathbf{D}_{-x}}$, i.e., $\mathbf{x}\in \{\mathcal{S}|\mathcal{S}=\mathcal{S}_{\mathbf{D}_{x}}\cap \mathcal{S}_{\mathbf{D}_{-x}}\}$, where $\mathcal{S}_{\mathbf{D}_{x}}$ and $\mathcal{S}_{\mathbf{D}_{-x}}$ denote the subspace spanned by $\mathbf{D}_{x}$ and $\mathbf{D}_{-x}$, respectively. The dimensionality of $\mathcal{S}_{\mathbf{D}_{x}}$ is $r_{x}$, and that of $\mathcal{S}_{\mathbf{D}_{-x}}$ is $r_{-x}$. Let $\mathbf{c}^{\ast}$ be the optimal solution of
\begin{equation}
\label{theo:equ3}
\min\hspace{1mm}\|\mathbf{c}\|_p \hspace{3mm} \mathrm{s.t.}\hspace{1mm} \mathbf{x}=\mathbf{D}\mathbf{c}
\end{equation}
over $\mathbf{D}=[\mathbf{D}_{x}\ \mathbf{D}_{-x}]$, and $\mathbf{c}^{\ast}=\begin{bmatrix}\mathbf{c}_{\mathbf{D}_{x}}^{\ast}\\\mathbf{c}_{\mathbf{D}_{-x}}^{\ast} \end{bmatrix}$ are partitioned according to the sets $\mathbf{D}_{x}$ and $\mathbf{D}_{-x}$. If
\begin{equation}
\label{theo:equ4}
\sigma_{\min}(\mathbf{D}_{x})\geq r_{-x} \cos\theta_{\min}\|\mathbf{D}_{-x}\|_{1,2},
\end{equation}
is satisfied, then $[\mathbf{c}_{\mathbf{D}_{x}}^{\ast}]_{r_{x},1}>[\mathbf{c}_{\mathbf{D}_{-x}}^{\ast}]_{1,1}$. Here, $\sigma_{\min}(\mathbf{D}_{x})$ is the smallest nonzero singular value of $\mathbf{D}_{x}$, $\theta_{\min}$ is the first principal angle between $\mathbf{D}_{x}$ and $\mathbf{D}_{-x}$, $\|\mathbf{D}_{-x}\|_{1,2}$ is the maximum $\ell_2$-norm of the columns of $\mathbf{D}_{-x}$, and $[\mathbf{c}]_{r,1}$ denotes the $r$-th largest absolute value of the entries of $\mathbf{c}$.
\end{theorem}

\begin{proof}
Since $\mathbf{x}\in \{\mathcal{S}|\mathcal{S}=\mathcal{S}_{\mathbf{D}_{x}}\cap \mathcal{S}_{\mathbf{D}_{-x}}\}$, we could write $\mathbf{x}=\mathbf{U}_{r_{0}} \mathbf{\Sigma}_{r_{0}} \mathbf{V}_{r_{0}}^{T}  \mathbf{z}_{\mathbf{D}_{x}}$, where $\mathbf{D}_{x}=\mathbf{U}_{r_{0}} \mathbf{\Sigma}_{r_{0}} \mathbf{V}_{r_{0}}^{T}$ is the skinny SVD of $\mathbf{D}_{x}$, $\mathbf{\Sigma}_{r_{0}}=\mathrm{diag}(\sigma_1(\mathbf{D}_{x}), \sigma_2(\mathbf{D}_{x}), \cdots, \sigma_{r_{0}}(\mathbf{D}_{x}))$, $r_0$ is the rank of $\mathbf{D}_{x}$, and $\mathbf{z}_{\mathbf{D}_{x}}$ is the optimal solution of (\ref{theo:equ3}) over $\mathbf{D}_{x}$. Thus, $\mathbf{z}_{\mathbf{D}_{x}}= \mathbf{V}_{r_{0}} \mathbf{\Sigma}_{r_{0}}^{-1} \mathbf{U}_{r_{0}}^{T} \mathbf{x}$.

From the propositions of $p$-norm, i.e., $\|\mathbf{z}\|_{\infty} \leq \|\mathbf{z}\|_{1} \leq n\|\mathbf{z}\|_{\infty}$, $\|\mathbf{z}\|_{\infty} \leq \|\mathbf{z}\|_{2} \leq \sqrt{n}\|\mathbf{z}\|_{\infty}$, and $\|\mathbf{z}\|_{2} \leq \|\mathbf{z}\|_{1} \leq \sqrt{n}\|\mathbf{z}\|_{2}$, we have
\begin{equation}
\label{lem3:equ1}
\|\mathbf{z}_{\mathbf{D}_{x}}\|_p
\leq \|\mathbf{z}_{\mathbf{D}_{x}}\|_1
\leq \sqrt{r_0}\|\mathbf{z}_{\mathbf{D}_{x}}\|_2
= \sqrt{r_0}\left\|\mathbf{V}_{r_{0}} \mathbf{\Sigma}_{r_{0}}^{-1} \mathbf{U}_{r_{0}}^{T} \mathbf{x}\right\|_2.
\end{equation}

Since the Frobenius norm is subordinate to the Euclidean vector norm, we must have

\begin{align}
\label{lem3:equ2}
    \|\mathbf{z}_{\mathbf{D}_{x}}\|_p
    &\leq \sqrt{r_0}\left\|\mathbf{V}_{r_{0}} \mathbf{\Sigma}_{r_{0}}^{-1} \mathbf{U}_{r_{0}}^{T}\|_{F}\|\mathbf{x}\right\|_2\notag\\
    &=\frac{\sqrt{r_0}}{\sqrt{\sigma_1^2(\mathbf{D}_{x}) + \cdots + \sigma_{r_0}^2(\mathbf{D}_{x})}}\|\mathbf{x}\|_2\notag\\
    &\leq \sigma_{\min}^{-1}(\mathbf{D}_{x})\|\mathbf{x}\|_2
\end{align}
where $\sigma_{\min}(\mathbf{D}_{x})=\sigma_{r_0}(\mathbf{D}_{x})$ is the smallest nonzero singular value of $\mathbf{D}_{x}$.

Moreover, $\mathbf{x}$ could be represented as a linear combination of $\mathbf{D}_{-x}$ since it lies in the intersection between $\mathcal{S}_{\mathbf{D}_{x}}$ and $\mathcal{S}_{\mathbf{D}_{-x}}$, i.e., $\mathbf{x}=\mathbf{D}_{-x}\mathbf{z}_{\mathbf{D}_{-x}}$, where $\mathbf{z}_{\mathbf{D}_{-x}}$ is the optimal solution of (\ref{theo:equ3}) over $\mathbf{D}_{-x}$. Multiplying two sides of the equation with $\mathbf{x}^T$, it gives $\|\mathbf{x}\|_2=\mathbf{x}^{T} \mathbf{D}_{-x} \mathbf{z}_{\mathbf{D}_{-x}}$. According to the H\"{o}lder's inequality, we have
\begin{equation}
\label{lem3:equ3}
\|\mathbf{x}\|_2^2
  \leq \|\mathbf{D}_{-x}^T \mathbf{x}\|_{\infty} \|\mathbf{z}_{\mathbf{D}_{-x}}\|_1,
\end{equation}

According to the definition of the first principal angles (Definition~\ref{def1}), we have

\begin{align}
\label{lem3:equ4}
\|\mathbf{D}_{-x}^T \mathbf{x}\|_{\infty}
  &=    \max\left(\left|[\mathbf{D}_{-x}]_{1}^{T}\mathbf{x}\right|, \left|[\mathbf{D}_{-x}]_{2}^{T}\mathbf{x}\right|, \cdots \right)\notag\\
  &\leq \cos\theta_{\min}\|\mathbf{D}_{-x}\|_{1,2}\|\mathbf{x}\|_2,
\end{align}
where $[\mathbf{D}_{-x}]_{i}$ denotes the $i$th column of $\mathbf{D}_{-x}$, $\theta_{\min}$ is the first principal angle between $\mathcal{S}_{\mathbf{D}_{x}}$ and $\mathcal{S}_{\mathbf{D}_{-x}}$, and $\|\mathbf{D}_{-x}\|_{1,2}$ denotes the maximum $\ell_2$-norm of the columns of $\mathbf{D}_{-x}$. Note that the smallest principal angle between any two subspaces always greater than zero, hence, $\cos\theta_{\min}\in[0,1)$.

Combining (\ref{lem3:equ3}) and (\ref{lem3:equ4}), it gives that
\begin{equation}
\label{lem3:equ5}
\|\mathbf{x}\|_2^2
  \leq \cos\theta_{\min}\|\mathbf{D}_{-x}\|_{1,2}\|\mathbf{x}\|_2 \|\mathbf{z}_{\mathbf{D}_{-x}}\|_1,
\end{equation}
hence,
\begin{equation}
\label{lem3:equ6}
\|\mathbf{z}_{\mathbf{D}_{-x}}\|_1
  \geq \frac{\|\mathbf{x}\|_2}{\cos\theta_{\min} [\mathbf{D}_{-x}]_{1,2}}.
\end{equation}

From the propositions of $p$-norm, we have
\begin{equation}
\label{lem3:equ7}
\|\mathbf{z}_{\mathbf{D}_{-x}}\|_p \geq \frac{\|\mathbf{x}\|_2}{\cos\theta_{\min} [\mathbf{D}_{-x}]_{1,2}}.
\end{equation}

Let $\|\mathbf{z}_{\mathbf{D}_{x}}\|_p < \|\mathbf{z}_{\mathbf{D}_{-x}}\|_p$, then
\begin{equation}
\label{lem3:equ8}
\sigma_{\min}^{-1}(\mathbf{D}_{x})\|\mathbf{x}\|_2 < \frac{\|\mathbf{x}\|_2}{\cos\theta_{\min} [\mathbf{D}_{-x}]_{1,2}},
\end{equation}
then,
\begin{equation}
\label{lem3:equ9}
\sigma_{\min}(\mathbf{D}_{x}) > \cos\theta_{\min} [\mathbf{D}_{-x}]_{1,2}.
\end{equation}

It is the sufficient condition for $[\mathbf{c}_{\mathbf{D}_{x}}^{\ast}]_{r_{0},1}>[\mathbf{c}_{\mathbf{D}_{-x}}^{\ast}]_{1,1}$ since it implies $\mathbf{c}_{\mathbf{D}_{x}}^{\ast}\neq \mathbf{0}$ and $\mathbf{c}_{\mathbf{D}_{-x}}^{\ast}= \mathbf{0}$ from Theorem~\ref{lem2}.

This completes the proof.
\end{proof}

\begin{figure*}[t]
\centering{
\subfigure[]{\label{fig1a}\includegraphics[width=0.2\textwidth]{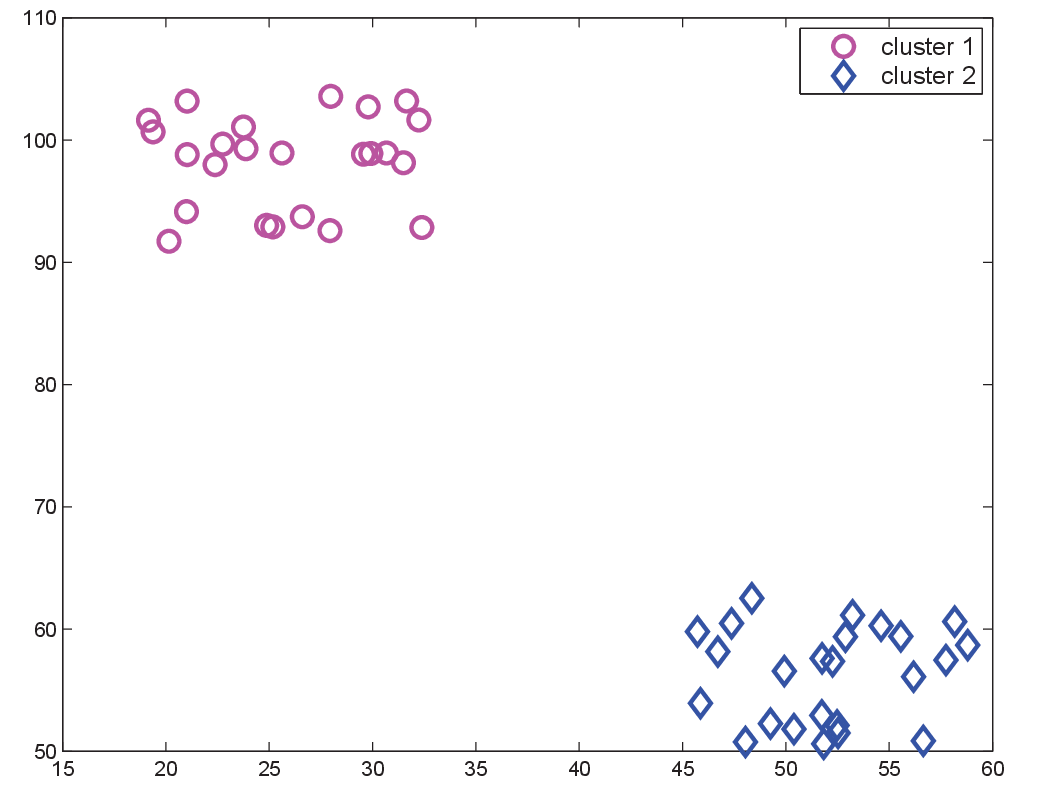}}
\subfigure[]{\label{fig1b}\includegraphics[width=0.18\textwidth]{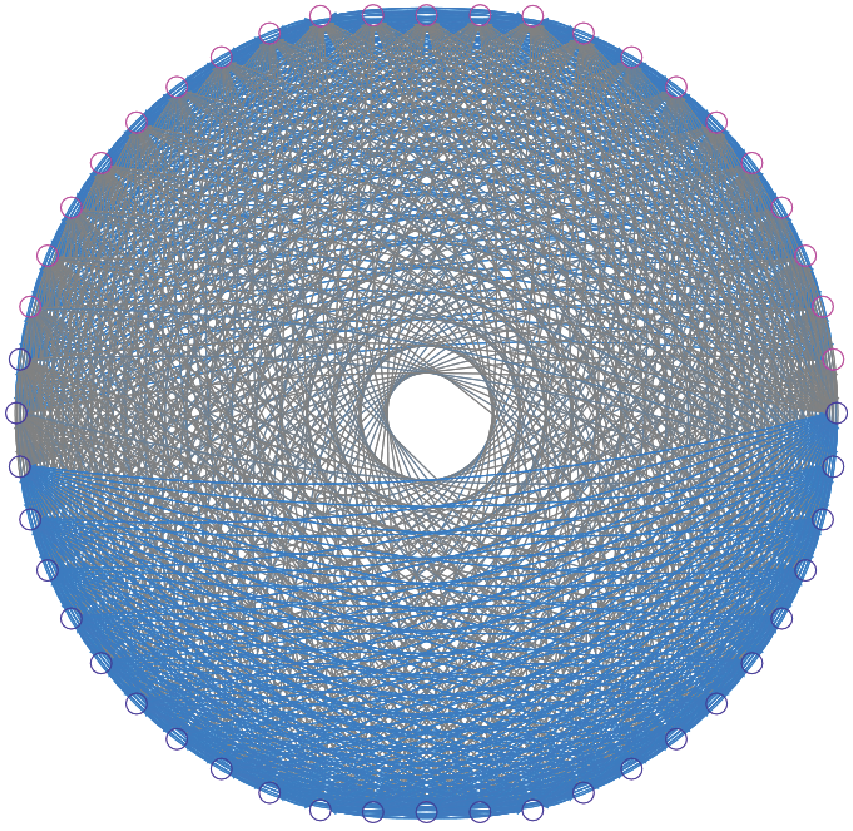}}
\subfigure[]{\label{fig:1.c}\includegraphics[width=0.2\textwidth]{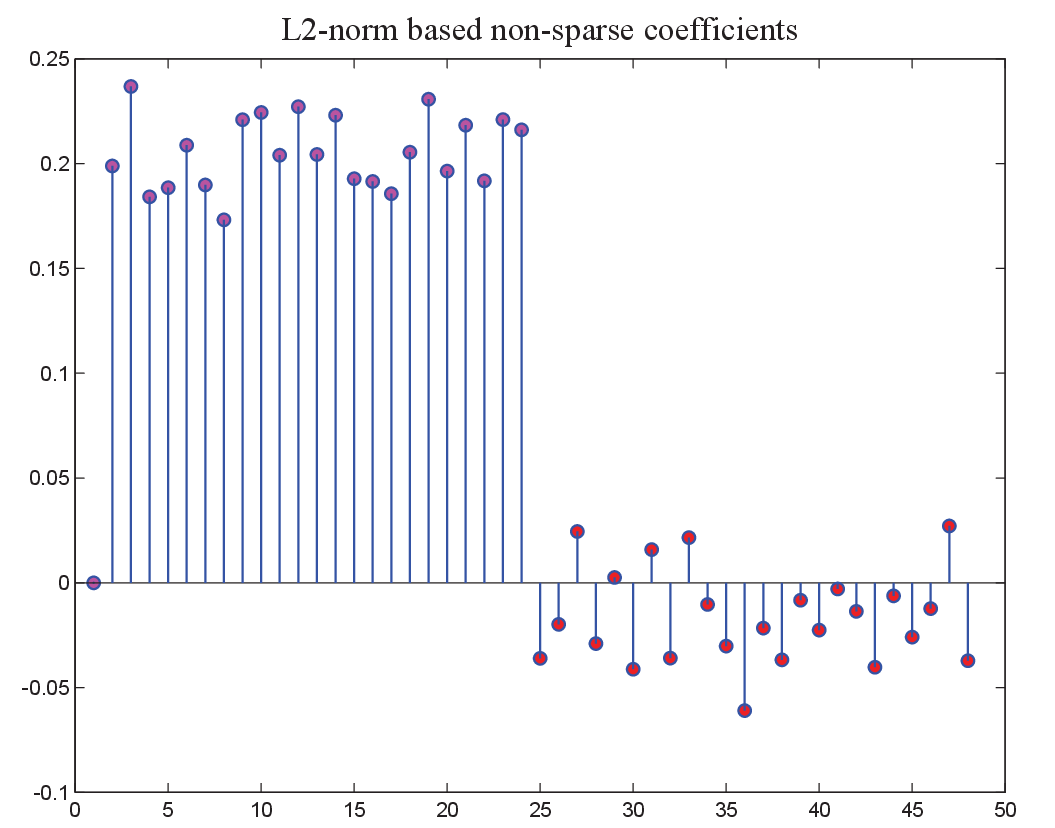}}
\subfigure[]{\label{fig1d}\includegraphics[width=0.18\textwidth]{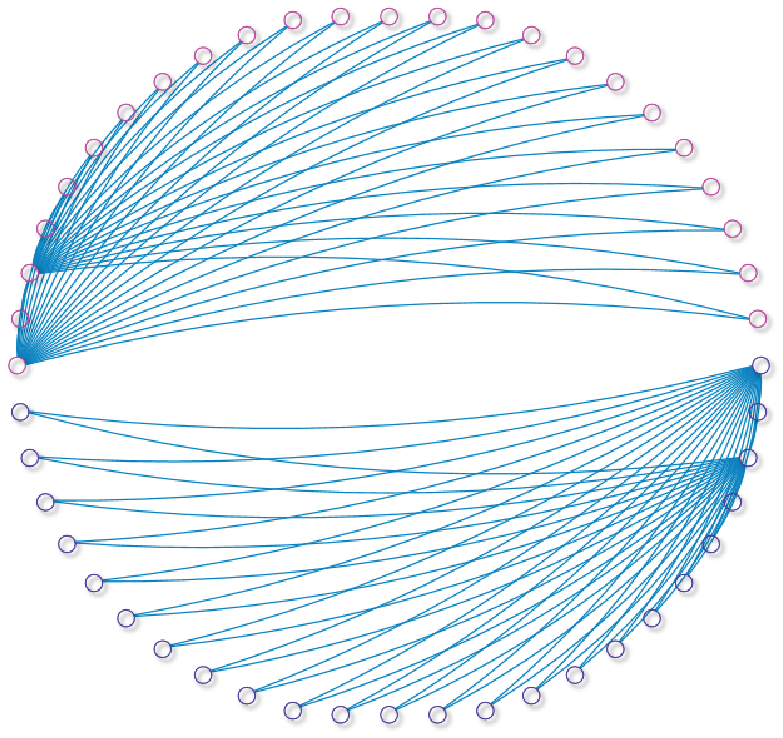}}
\subfigure[]{\label{fig1e}\includegraphics[width=0.2\textwidth]{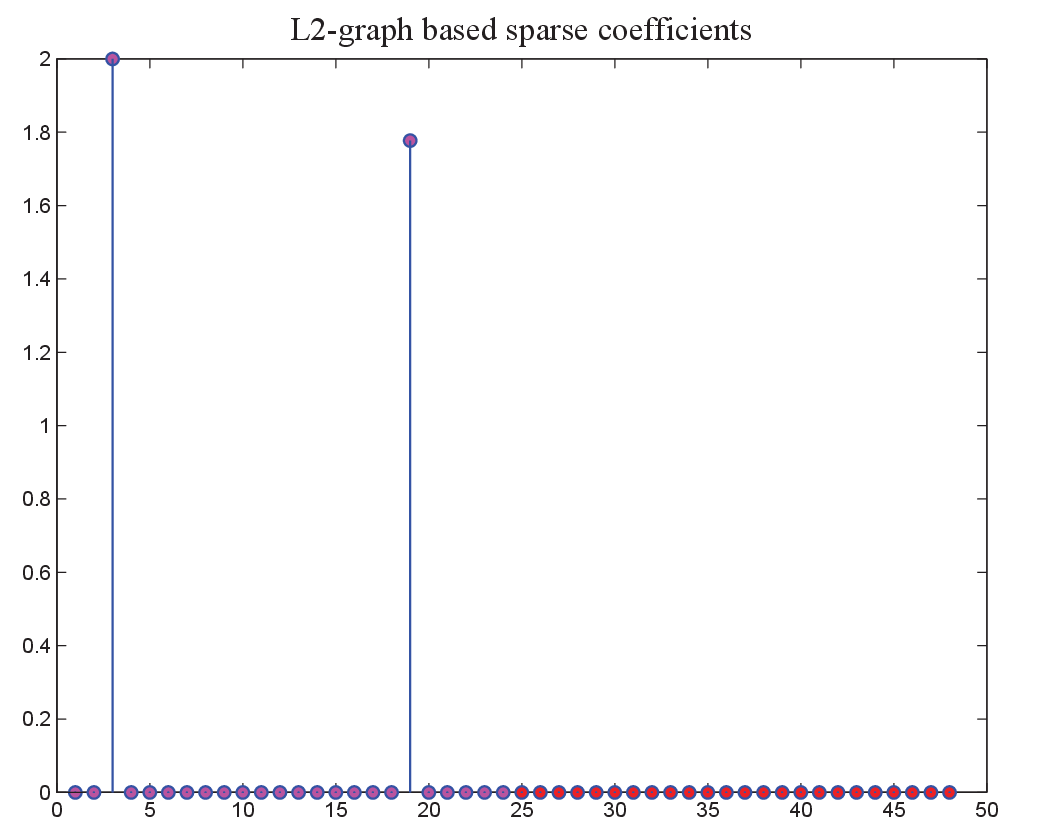}}}
\caption{A toy example of the intra-subspace projection dominance in $\ell_2$-norm-based projection space.
(a) A given data sets come from two clusters, indicated by different shapes. Note that each cluster corresponds to a subspace, and the two subspaces are dependent.
(b, c) The similarity graph in $\ell_2$-norm-based projection space and the coefficients of a data point $\mathbf{x}$. The first and the last 25 values in (c) correspond to the coefficients (similarity) over the intra-cluster and inter-cluster data points, respectively.
(d, e) The similarity graph achieved by our method and the coefficients of $\mathbf{x}$. For each data point, only the $2$ largest coefficients are nonzero, corresponding to the projection over the base of $\mathds{R}^{2}$.
From (b) and (d), the inter-cluster data points connections are removed and the data are successfully separated into respective clusters. }
\label{fig1}
\end{figure*}

\subsection{IPD in Nuclear-norm based Projection Space}
\label{sec3.2}

Nuclear-norm has been widely used as a convex relaxation of rank-minimization problem. Based on two theoretical results~\cite{Liu2013,Favaro2011}, we show that the IPD property is also shared by the nuclear-norm case.

\begin{lemma}[\cite{Liu2013}]
\label{lem4}
Let $\mathbf{D}=\mathbf{U}_{r} \mathbf{\Sigma}_{r} \mathbf{V}_{r}^{T}$ be the skinny singular value decomposition (SVD) of the data matrix $\mathbf{D}$. The unique solution to
\begin{equation}
\label{lem4:equ1}
    \min\hspace{1mm}\|\mathbf{C}\|_{\ast} \hspace{3mm}
    \mathrm{s.t.} \hspace{1mm} \mathbf{D} = \mathbf{D} \mathbf{C}
\end{equation}
is given by $\mathbf{C}^{\ast} = \mathbf{V}_{r}\mathbf{V}_{r}^{T}$, where $r$ is the rank of $\mathbf{D}$.
\end{lemma}

Note that, Lemma~\ref{lem4} implies the assumption that the data matrix $\mathbf{D}$ is free to errors.

\begin{lemma}[\cite{Favaro2011}]
\label{lem5}
Let $\mathbf{D}=\mathbf{U} \mathbf{\Sigma} \mathbf{V}^{T}$ be the SVD of the data matrix $\mathbf{D}$. The optimal solution to
\begin{equation}
\label{lem5:equ1}
\underset{\mathbf{C},\mathbf{D}_{0}}{\min}\hspace{1mm} \|\mathbf{C}\|_{\ast}+\frac{\alpha}{2}\|\mathbf{D} - \mathbf{D}_{0}\|_{F}^{2} \hspace{3mm}
\mathrm{s.t.} \hspace{1mm} \mathbf{D}_{0}=\mathbf{D}_{0}\mathbf{C}
\end{equation}
is given by $\mathbf{D}_{0}^{\ast}=\mathbf{U}_1 \mathbf{\Sigma}_1 \mathbf{V}_{1}^{T}$ and $\mathbf{C}^{\ast}=\mathbf{V}_{1} \mathbf{V}_{1}^{T}$, where $\mathbf{\Sigma}_1$, $\mathbf{U}_1$ and $\mathbf{V}_1$ are the top $k^{\ast}=\underset{k}{\mathrm{argmin}}\hspace{1mm}k+\frac{\alpha}{2} \underset{i>k}{\sum}\sigma_{i}^{2}$ singular values and singular vectors of $\mathbf{D}$, respectively.
\end{lemma}

\begin{theorem}
\label{lem6}

Let $\mathbf{C}^{\ast}=\mathbf{U}_{\mathbf{C}} \mathbf{\Sigma}_{\mathbf{C}} \mathbf{V}_{\mathbf{C}}^{T}$ be the skinny SVD of the optimal solution to
\begin{equation}
\label{lem6:equ1}
\min \hspace{1mm} \|\mathbf{C}\|_{\ast} \hspace{3mm}
\mathrm{s.t.} \hspace{1mm} \mathbf{D}=\mathbf{D}\mathbf{C},
\end{equation}
where $\mathbf{D}$ consists of the clean data set $\mathbf{D}_0$ and the errors $\mathbf{D}_e$, i.e., $\mathbf{D}=\mathbf{D}_{0}+\mathbf{D}_{e}$.

The optimal solution to
\begin{equation}
\label{lem6:equ2}
\underset{\mathbf{C}_0,\mathbf{D}_0}{\min}\hspace{1mm} \|\mathbf{C}_0\|_{\ast}+\frac{\alpha}{2}\|\mathbf{D}_e\|_{F}^{2} \hspace{3mm}
\mathrm{s.t.} \hspace{1mm} \mathbf{D}_0=\mathbf{D}_0\mathbf{C}_0, \mathbf{D} = \mathbf{D}_0 + \mathbf{D}_e
\end{equation}
is given by $\mathbf{C}_{0}^{\ast}=\mathbf{U}_{\mathbf{C}}\mathcal{H}_{k^{\ast}}(\mathbf{\Sigma}_{\mathbf{C}}) \mathbf{V}_{\mathbf{C}}^{T}$, where $\mathcal{H}_{k}(\mathbf{x})$ is a truncation operator that retains the first $k$ elements and sets the other elements to zero, $k^{\ast}=\underset{k}{\mathrm{argmin}}\hspace{1mm}k+\frac{\alpha}{2} \underset{i>k}{\sum}\sigma_{i}^{2}$, and $\sigma_{i}$ is the $i$th largest singular value of $\mathbf{D}$.
\end{theorem}

\begin{proof}
Suppose the rank of data matrix $\mathbf{D}$ is $r$, let $\mathbf{D}=\mathbf{U} \mathbf{\Sigma} \mathbf{V}^{T}$ and $\mathbf{D}=\mathbf{U}_{r} \mathbf{\Sigma}_{r} \mathbf{V}_{r}^{T}$ be the SVD and skinny SVD of $\mathbf{D}$, respectively. Hence, we have $\mathbf{U}= [\mathbf{U}_{r}\ \mathbf{U}_{-r}]$, $\mathbf{\Sigma} = \begin{bmatrix}\mathbf{\Sigma}_{r} & \mathbf{0}\\ \mathbf{0} & \mathbf{0} \end{bmatrix}$ and $\mathbf{V} = \begin{bmatrix} \mathbf{V}_{r}^{T}\\ \mathbf{V}_{-r}^{T} \end{bmatrix}$, where $\mathbf{I}=\mathbf{U}_{r}^{T}\mathbf{U}_{r}+\mathbf{U}_{-r}^{T}\mathbf{U}_{-r}$, $\mathbf{I}=\mathbf{V}_{r}^{T}\mathbf{V}_{r}+\mathbf{V}_{-r}^{T}\mathbf{V}_{-r}$, $\mathbf{U}_{r}^{T}\mathbf{U}_{-r}=\mathbf{0}$, and $\mathbf{V}_{r}^{T}\mathbf{V}_{-r}=\mathbf{0}$.

On the one hand, from Lemma~\ref{lem4}, the optimal solution of (\ref{lem6:equ1}) is given by $\mathbf{C}^{\ast}=\mathbf{V}_{r}\mathbf{V}_{r}^{T}$ which is a solid skinny SVD for $\mathbf{C}^{\ast}$. Therefore, we can choose $\mathbf{U}_\mathbf{C} = \mathbf{V}_{r}$, $\mathbf{\Sigma}_\mathbf{C} = \mathbf{I}$ and $\mathbf{V}_\mathbf{C} = \mathbf{V}_{r}$.

On the other hand, from Lemma~\ref{lem5}, the optimal solution of (\ref{lem6:equ2}) is given by $\mathbf{C}_{0}^{\ast}=\mathbf{V}_{1}\mathbf{V}_{1}^{T}$, where $\mathbf{V}_1$ is the top $k^{\ast}=\underset{k}{\mathrm{argmin}}\hspace{1mm}k+\frac{\alpha}{2} \underset{i>k}{\sum}\sigma_{i}^{2}$ right singular vectors of $\mathbf{D}$. Therefore, we can conclude that $\mathbf{V}_1$ corresponds to the top $k^{\ast}$ singular vector of $\mathbf{V}_r$ owing to $k^{\ast}\leq r$, i.e., $\mathbf{C}_{0}^{\ast}=\mathbf{U}_{\mathbf{C}}\mathcal{H}_{k^{\ast}}(\mathbf{\Sigma}_{\mathbf{C}}) \mathbf{V}_{\mathbf{C}}^{T}$, where $\mathcal{H}_{k}(\mathbf{x})$ keeps the first $k$ elements and sets the other elements to zero.

This completes the proof.
\end{proof}

The IPD property forms the fundamental theoretical basis for the subsequent L2-Graph algorithm. According to the IPD, the coefficients over intra-subspace is always larger than those over the errors in terms of $\ell_p$- and nuclear-norm based projection space. Hence, the effect of the errors can be eliminated by keeping $k$ largest entries and zeroing the other entries, where $k$ equals to the dimensionality of the corresponding subspace. We summarize such errors-handling method as `\textbf{encoding and then removing errors from projection space}'. Compared with the popular method  `removing errors from input space and then encoding', the proposed method no longer requires the prior knowledge on the structure of errors.

\figurename~\ref{fig1} shows a toy example  illustrating the intra-subspace projection dominance in the $\ell_2$-norm-based projection space, where the data points are sampled from two dependent subspaces corresponding to two clusters in $\mathds{R}^{2}$. In this example, the errors (the intersection between two dependent subspaces) lead to the connections between the inter-cluster data points and the weights of these connections are smaller than the edge weights between the intra-cluster data points (\figurename~\ref{fig1b}). By thresholding the connections with trivial weight, we obtain a new similarity graph as shown in (\figurename~\ref{fig1d}). Clearly, this toy example again shows the IPD property of $\ell_2$-norm-based projection space and the effectiveness of the proposed errors-removing method.

\section{Constructing the L2-Graph for Robust Subspace Learning and Subspace Clustering}
\label{sec:4}

In this section, we present the L2-Graph method based on the IPD property of $\ell_2$-norm based projection space. We chose $\ell_2$-norm rather than the others such as $\ell_1$-norm since $\ell_2$-norm based objective function can be analytically solved. Moreover, we generalize our proposed framework to subspace clustering and subspace learning by incorporating L2-Graph into spectral clustering~\cite{Ng2002} and subspace learning~\cite{He2005}.

\subsection{Algorithms Description}
\label{sec:3.2}

Let $\mathbf{X} = \{\mathbf{x} _1,\mathbf{x}_2,\ldots,\mathbf{x}_n\}$ be a collection of data points located on a union of dependent or disjoint or independent subspaces $\{S_1, S_2, \ldots, S_L\}$ and $\mathbf{X}_i = [\mathbf{x}_1, \ldots, \mathbf{x}_{i-1},
\mathbf{0}, \mathbf{x}_{i+1}, \ldots, \mathbf{x}_n], (i=1,\cdots,n)$ be the dictionary for $\mathbf{x}_i$, we aim to solve the following problem:
\begin{equation}
\label{sec3.2.1} \mathop{\mathrm{min}}_{\mathbf{c}_i}
\hspace{1mm}\frac{1}{2}\|\mathbf{x}_i -
\mathbf{X}_i\mathbf{c}_i\|_2^2+\lambda\|\mathbf{c}_i\|_2^2,
\end{equation}
where $\lambda$ is a positive real number.

Equation (\ref{sec3.2.1}) is actually the well known ridge regression problem~\cite{Hoerl1970}, which has been investigated in the context of face recognition~\cite{Zhang2011}. There is, however, a lack of examination on its performance in subspace clustering and subspace learning. The optimal solution of (\ref{sec3.2.1}) is $(\mathbf{X}_i^T\mathbf{X}_i+\lambda\mathbf{I})^{-1}\mathbf{X}_i^{T}\mathbf{x}_i$ whose computational complexity is  $O(mn^4)$ for given $n$ data points with $m$ dimensions. To solve (\ref{sec3.2.1}) efficiently, we rewrite it as
\begin{equation}
\label{sec3.2.2}
\mathop{\min}_{\mathbf{c}_i}
\hspace{1mm}\frac{1}{2}\|\mathbf{x}_i -
\mathbf{X}\mathbf{c}_i\|_2^2+\lambda\|\mathbf{c}_i\|_2^2,
\hspace{2mm}
\mathrm{s.t.} \hspace{1mm} \mathbf{e}_{i}^T
\mathbf{c}_{i} = 0.
\end{equation}
Using Lagrangian method, we have
\begin{equation}
\label{sec3.2.3}
 \mathds{L}(\mathbf{c}_i) = \frac{1}{2}\|\mathbf{x}_i
- \mathbf{X}\mathbf{c}_i\|_2^2+\lambda\|\mathbf{c}_i\|_2^2 +
\gamma \mathbf{e}_{i}^T \mathbf{c}_{i},
\end{equation}
where $\gamma$ is the Lagrangian multiplier. Clearly,
\begin{equation}
\label{sec3.2.4}
\frac{\partial{\mathds{L}(\mathbf{c}_i)}}{\partial{\mathbf{c}_i}}=\left(\mathbf{X}^T\mathbf{X}+\lambda
\mathbf{I}\right) \mathbf{c}_i - \mathbf{X}^T\mathbf{x}_i+\gamma\mathbf{e}_i .
\end{equation}

Let $\frac{\partial{\mathds{L}(\mathbf{c}_i)}}{\partial{\mathbf{c}_i}}=0$, we obtain
\begin{equation}
\label{sec3.2.5}
\mathbf{c}_i = \left(\mathbf{X}^T \mathbf{X} + \lambda
\mathbf{I}\right)^{-1} \left(\mathbf{X}^T \mathbf{x}_i - \gamma
\mathbf{e}_i\right).
\end{equation}

Multiplying both sides of (\ref{sec3.2.5}) by $\mathbf{e}_i^T$, and since $\mathbf{e}_i^T\mathbf{c}_i =0$, it holds that
\begin{equation}
\label{sec3.2.6}
\gamma = \frac{\mathbf{e}_{i}^{T} \left(\mathbf{X}^{T}\mathbf{X} +
\lambda \mathbf{I}\right)^{-1} \mathbf{X}^{T}
\mathbf{x}_{i}}{\mathbf{e}_{i}^{T} \left(\mathbf{X}^{T}\mathbf{X}
+ \lambda \mathbf{I}\right)^{-1} \mathbf{e}_{i}}.
\end{equation}

Substituting $\gamma$ into (\ref{sec3.2.6}), the optimal solution is given by
\begin{equation}
\label{sec3.2.7}
\mathbf{c}_i^{\ast} = \mathbf{P}
\left[\mathbf{X}^{T}
\mathbf{x}_{i} -
\frac{\mathbf{e}_{i}^{T}\mathbf{Q} \mathbf{x}_{i}\mathbf{e}_{i}}
{\mathbf{e}_{i}^{T}\mathbf{P}\mathbf{e}_{i}}
\right],
\end{equation}
where $\mathbf{Q}=\mathbf{P}\mathbf{X}^{T}$, $\mathbf{P}=\left(\mathbf{D}^{T}\mathbf{D}
+ \lambda \mathbf{I}\right)^{-1}$, and the union of $\mathbf{e}_i$ $(i=1, \cdots, n)$ is the standard orthogonal basis of $\mathds{R}^n$, i.e., all entries in $\mathbf{e}_i$ are zeroes except the $i$-th entry is one.

After projecting the data set into the linear space spanned by itself via (\ref{sec3.2.7}), L2-Graph handles the errors by performing a hard thresholding operator $\mathcal{H}_{k}(\cdot)$ over $\mathbf{c}_i$, where $\mathcal{H}_{k}(\cdot)$ keeps $k$ largest entries in $\mathbf{c}_i$ and zeroing the others. Generally, the optimal $k$ equals to the dimensionality of corresponding subspace.

Once the L2-Graph was built, we perform subspace learning and subspace clustering with it. The proposed methods are summarized in Algorithms~\ref{algorithm1} and \ref{algorithm2}.

\begin{algorithm}[t]
    \caption{Robust Subspace Learning with L2-Graph}
    \label{algorithm1}
    \begin{algorithmic}[1]
    \REQUIRE
        A given data set $\mathbf{X}=\{\mathbf{x}_i\}_{i=1}^n$, a new coming datum $\mathbf{y}\in span\{\mathbf{X}\}$, the balanced parameter $\lambda$ and the thresholding parameter $k$.
    \STATE Calculate $\mathbf{P}=\left(\mathbf{X}^{T}\mathbf{X} + \lambda \mathbf{I}\right)^{-1}$ and $\mathbf{Q}=\mathbf{PX}^{T}$ and store them.
    \STATE For each point $\mathbf{x}_{i}$, obtain its representation $\mathbf{c}_i$ via
        \begin{equation}
        \label{alg:equ1.1}
        \mathbf{c}_i^{\ast} = \mathbf{P}
        \left[\mathbf{X}^{T}
        \mathbf{x}_{i}-\frac{\mathbf{e}_{i}^{T}\mathbf{Q} \mathbf{x}_{i}\mathbf{e}_{i}}{\mathbf{e}_{i}^{T}\mathbf{P}\mathbf{e}_{i}}\right],
    \end{equation}
    \STATE  For each $\mathbf{c}_i$, eliminate the effect of errors in the projection space via $\mathbf{c}_i=\mathcal{H}_{k}(\mathbf{c}_i)$, where the hard thresholding operator $\mathcal{H}_{k}(\mathbf{c}_i)$ keeps $k$ largest entries in $\mathbf{c}_i$ and zeroes the others.
    \STATE Construct an affinity matrix by $\mathbf{W}_{ij}=|\mathbf{c}_{ij}|+|\mathbf{c}_{ji}|$ and normalize each column of $\mathbf{W}$ to have a unit $\ell_2$-norm, where $\mathbf{c}_{ij}$ is the $j$th entry of $\mathbf{c}_i$.
    \STATE Embed $\mathbf{W}$ into a $m^{\prime}$-dimensional space and calculate the projection matrix $\mathbf{\Theta}\in \mathds{R}^{m\times m^{\prime}}$ via solving
         \begin{equation}
         \label{alg:equ1.2}
             \mathop{\min}_{\mathbf{\Theta}}  \left\| \mathbf{\Theta} ^T\mathbf{D} - \mathbf{\Theta}^T  \mathbf{D} \mathbf{W} \right\|_F^2,
             \mathrm{\hspace{2mm}  s.t. \hspace{1mm} \mathbf{\Theta}^T\mathbf{D}\mathbf{D}^T  \mathbf{\Theta}=\mathbf{I}},
             \end{equation}
    \ENSURE The projection matrix $\mathbf{\Theta}$ and the low-dimensional representation of $\mathbf{y}$ via $\mathbf{z}=\mathbf{\Theta}^{T}\mathbf{y}$.
    \end{algorithmic}
\end{algorithm}

\begin{algorithm}[t]
    \caption{Robust Subspace Clustering with L2-Graph}
    \label{algorithm2}
    \begin{algorithmic}[1]
    \REQUIRE
    A collection of data points $\mathbf{X}=\{\mathbf{x}_i\}_{i=1}^n$ sampled from a union of linear subspaces $\{S_i\}_{i=1}^{c}$, the balance parameter $\lambda$ and thresholding parameter $k$;
    \STATE Calculate $\mathbf{P}=\left(\mathbf{X}^{T}\mathbf{X} + \lambda \mathbf{I}\right)^{-1}$ and $\mathbf{Q}=\mathbf{PX}^{T}$ and store them.
    \STATE For each point $\mathbf{x}_{i}$, obtain its representation $\mathbf{c}_i$ via
        \begin{equation}
        \label{alg:equ2.1}
        \mathbf{c}_i^{\ast} = \mathbf{P}
        \left[\mathbf{X}^{T}
        \mathbf{x}_{i}-\frac{\mathbf{e}_{i}^{T}\mathbf{Q} \mathbf{x}_{i}\mathbf{e}_{i}}{\mathbf{e}_{i}^{T}\mathbf{P}\mathbf{e}_{i}}\right],
    \end{equation}
    \STATE  For each $\mathbf{c}_i$, eliminate the effect of errors in the projection space via $\mathbf{c}_i=\mathcal{H}_{k}(\mathbf{c}_i)$, where the hard thresholding operator $\mathcal{H}_{k}(\mathbf{c}_i)$ keeps $k$ largest entries in $\mathbf{c}_i$ and zeroes the others.
    \STATE Construct an affinity matrix by $\mathbf{W}_{ij}=|\mathbf{c}_{ij}|+|\mathbf{c}_{ji}|$ and normalize each column of $\mathbf{W}$ to have a unit $\ell_2$-norm, where $\mathbf{c}_{ij}$ is the $j$th entry of $\mathbf{c}_i$.
    \STATE Construct a Laplacian matrix $\mathbf{L}=\mathbf{\Sigma}^{-1/2} \mathbf{W} \mathbf{\Sigma}^{-1/2}$, where $\mathbf{\Sigma}=\mathrm{diag}\{\sigma_{i}\}$ with $\sigma_{i}=\sum_{j=1}^{n}\mathbf{W}_{ij}$.
    \STATE Obtain the eigenvector matrix $\mathbf{V}\in \mathds{R}^{n\times c}$ which consists of the first $c$ normalized eigenvectors of $\mathbf{L}$ corresponding to its $c$ smallest nonzero eigenvalues.
    \STATE Perform k-means clustering algorithm on the rows of $\mathbf{V}$.
    \ENSURE The cluster assignment of $\mathbf{X}$.
    \end{algorithmic}
\end{algorithm}

\begin{figure*}[t]
\centering{
\subfigure[]{\label{fig2a}\centering\includegraphics[width=0.36\textwidth]{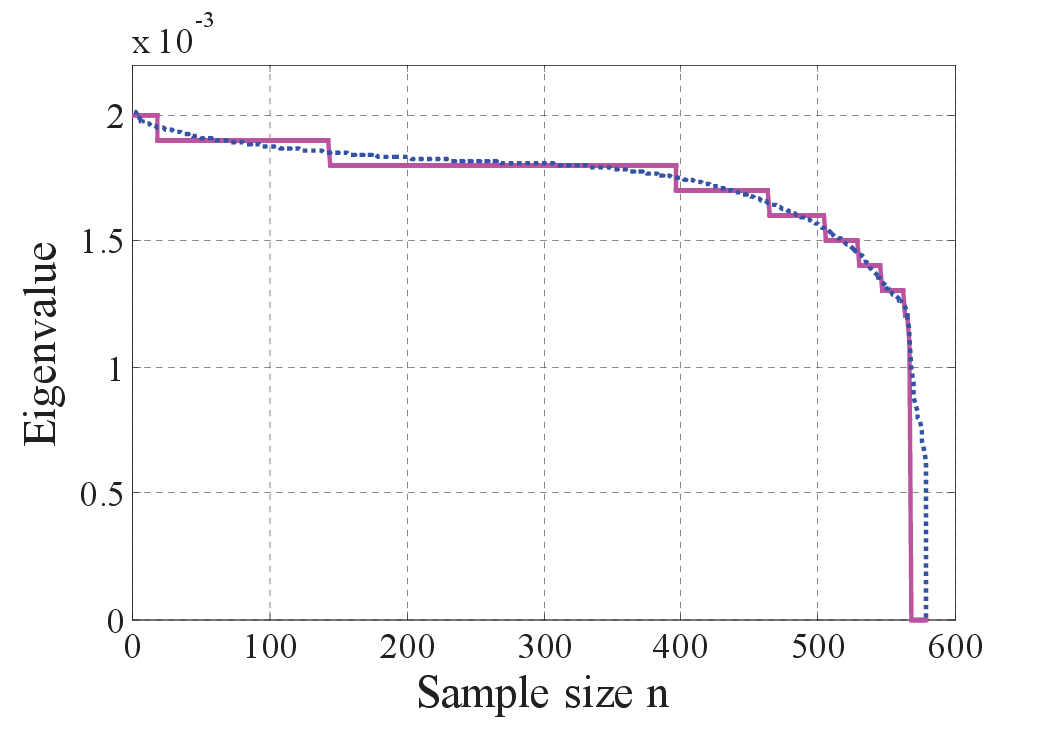}}
\subfigure[]{\label{fig2b}\centering\includegraphics[width=0.26\textwidth]{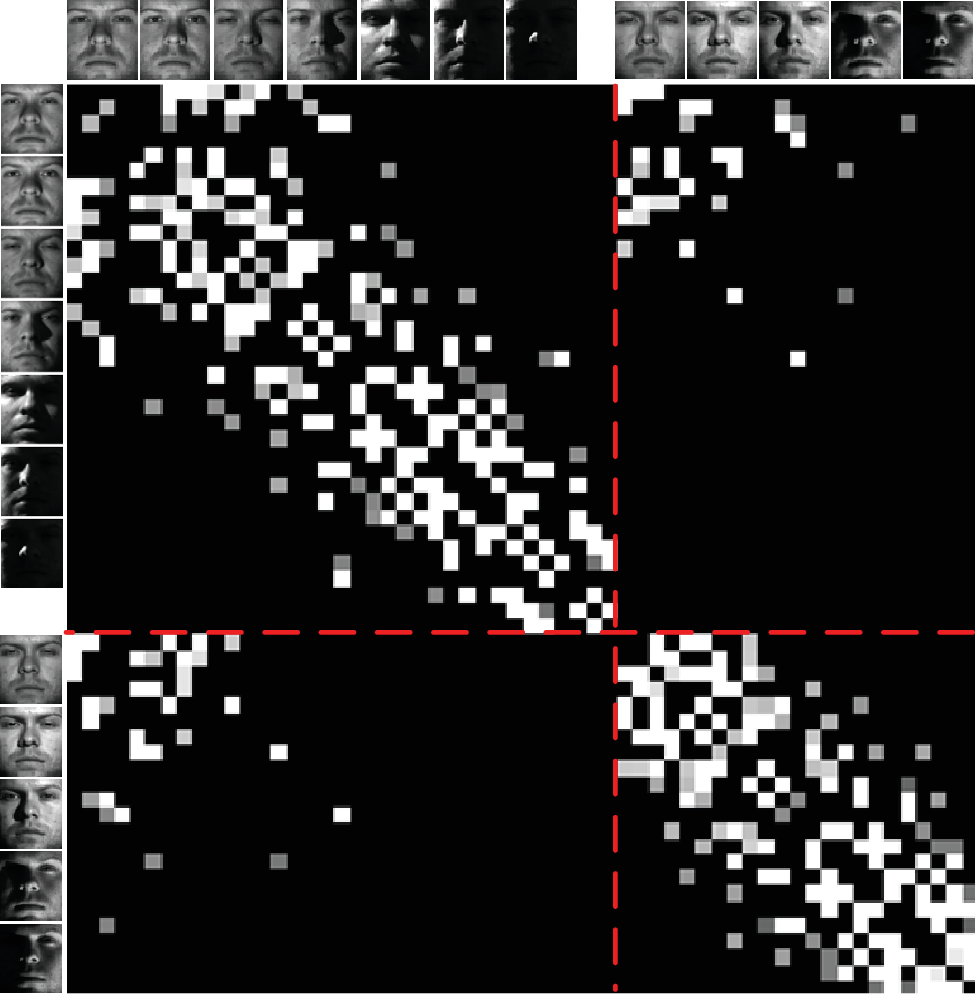}}
\subfigure[]{\label{fig2c}\centering\includegraphics[width=0.36\textwidth]{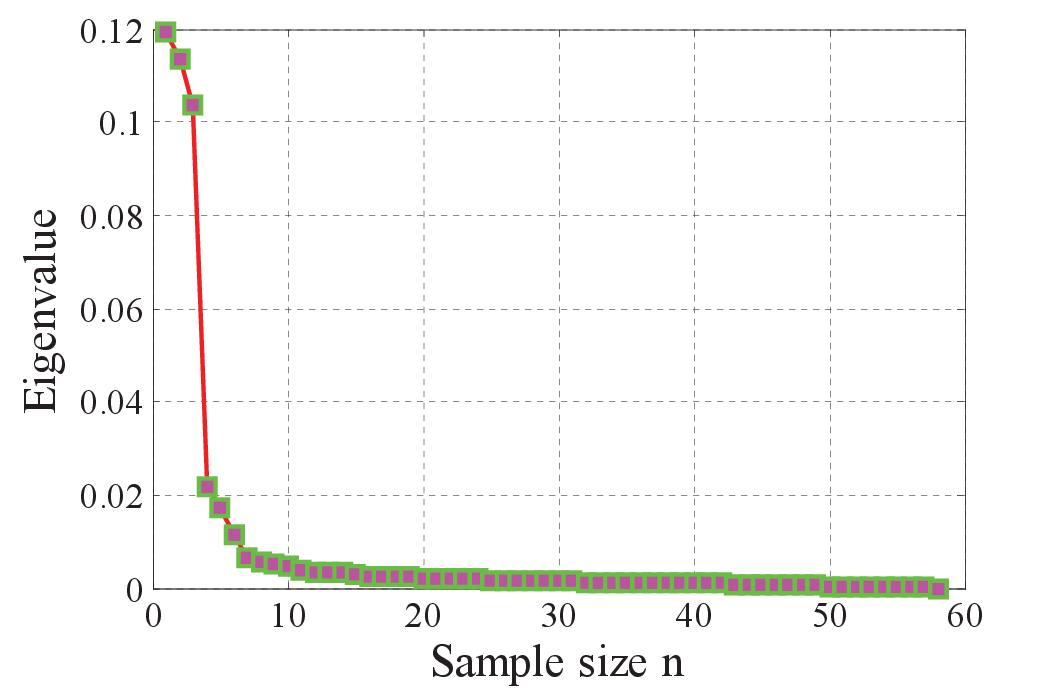}}
}
\caption{\label{fig2} Estimating the latent structure of a given data set. The used data set contains 580 frontal images drawn from the first 10 subjects of the Extended Yale B (ExYaleB)~\cite{Georghiades2001}.
(a) The dotted curve plots the eigenvalues of $\mathbf{L}$, and the red solid line plots the discretized eigenvalues. Clearly, the number of the unique nonzero eigenvalues is 10. This means the data set contains 10 subjects matching with the ground truth.
(b) The affinity matrix $\mathbf{W}\in \mathds{R}^{58\times 58}$ obtained by our algorithm. The experiment was carried out on the first 58 samples of the first subject of ExYaleB. The left column and the top row illustrate some images. The dotted lines split the matrix into four parts. The upper-left part: the similarity relationship among the first 32 images which are illuminated from right side. The bottom-right part: the relationship among the remaining 26 images which are illuminated from left side. From the connections, it is easy to find that our method reflects the variation in the direction of light source.
(c) The eigenvalues of $\mathbf{W}$. One could find that most energy concentrates to the first 6 components. This means that the intrinsic dimensionality of these data is around 6. The result is consistent with~\cite{Costa2004}.}
\end{figure*}

\subsection{Computational Complexity Analysis}
\label{sec:3.2}

Suppose the data points $\mathbf{X} \in \mathds{R}^{m\times n}$ are drawn from a union of subspaces. The L2-Graph takes $O(mn^2+n^3)$ to compute and store the matrices $\mathbf{P}=(\mathbf{X}^T\mathbf{X}+\lambda \mathbf{I})^{-1}$ and $\mathbf{Q}=\mathbf{P}\mathbf{X}^{T}$. It then projects each data point into another space via (\ref{sec3.2.7}) with complexity $O(mn)$. Moreover, to eliminate the effects of errors, it requires $O(k\log{k})$ to find $k$ largest coefficients. Putting everything together, the computational complexity of L2-Graph is $O(mn^2+n^3)$. This cost is considerably less than sparse representation based methods~\cite{Elhamifar2013,Qiao2010,Cheng2010}
($O(tm^2n^2+tmn^3)$) and low rank representation~\cite{Liu2013} ($O(tnm^2+tn^3)$), where $t$ denotes the iterative number of the corresponding optimizer.

\subsection{Estimating the Structure of Data Space with L2-Graph}
\label{sec:3.2}

In this section, we show how to estimate the number of subspaces, the sub-manifold of the given data set, and the subspace dimensionality  with L2-Graph.

When the obtained affinity matrix $\mathbf{W}$ is strictly block-diagonal, i.e., $\mathbf{W}_{ij}\neq 0$ if and only if the data points $\mathbf{d}_{i}$ and $\mathbf{d}_j$ belong to the same subspace, one can predict the number of subspace by counting the number of unique singular value of the Laplacian matrix $\mathbf{L}$ as suggested by~\cite{Luxburg2007}, where $\mathbf{L}=\mathbf{I}-\mathbf{\Sigma}^{-1/2}\mathbf{W}\mathbf{\Sigma}^{-1/2}$ and $\mathbf{\Sigma}=\mathrm{diag}(\sigma_{i})$ with $\sigma_{i}=\sum_{j=1}^{n}\mathbf{W}_{ij}$. In most cases, however, $\mathbf{W}$ is not strictly block-diagonal and therefore the method fails to get the correct result. \figurename~\ref{fig2a} shows an example by plotting the singular values of $\mathbf{L}$ derived upon L2-Graph (dotted curve). To solve this problem, we perform the DBSCAN method~\cite{Ester1996} to discretize the eigenvalues of $\mathbf{L}$. The processed singular values are plotted in the solid line. One can find that the values decrease from $0.02$ to $0.011$ with an interval of $0.001$. This shows that the number of subspace is $10$ and the result is in accordance  with the ground truth.

To estimate the intrinsic dimensionality of subspace, we give an example by using the first 58 samples from the first subject of Extended Yale database B and building an affinity matrix $\mathbf{W}$ using L2-Graph as shown in \figurename~\ref{fig2b}. We perform Principle Component Analysis (PCA) on $\mathbf{W}$ and count the number of the eigenvalues above a specified threshold. The number is regarded as  the intrinsic dimensionality of the subspace as shown in \figurename~\ref{fig2c}. Note that, \figurename~\ref{fig2b} shows that L2-Graph can also reveal the sub-manifold of the given data set, i.e., two sub-manifolds corresponding to two directions of light source in this example. This ability is helpful in understanding the latent data structure.

\section{Experimental Verification and Analysis}
\label{sec:5}

In this section, we evaluate the performance of the L2-Graph in the context of subspace learning and subspace clustering. Besides face clustering, we investigate the result of L2-Graph for another application of subspace clustering, i.e., motion segmentation. We consider the results in terms of three aspects: 1) accuracy, 2) robustness, and 3) computational cost.

\begin{figure*}[t]
\begin{center}
\subfigure []{\label{fig:3a}\includegraphics[width=0.46\textwidth]{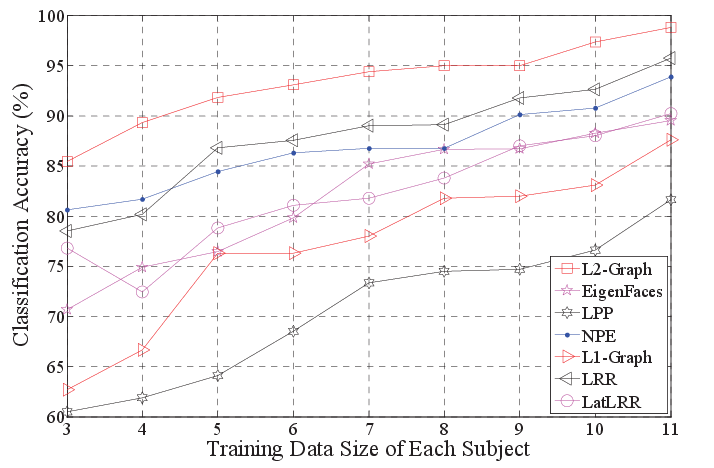}}
\subfigure []{\label{fig:3b}\includegraphics[width=0.46\textwidth]{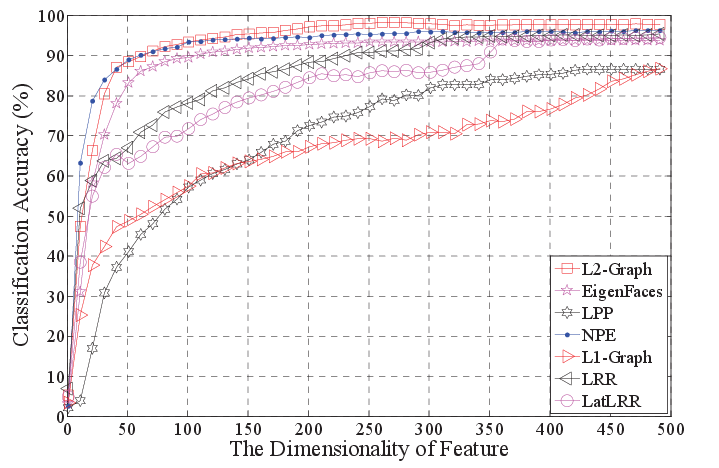}}
\end{center}
\caption{\label{fig:3} (a) The classification accuracy of the tested methods with increasing training AR1 images.  (b) The recognition rate of 1-NN classifier with different subspace learning methods over ExYaleB. }
\end{figure*}

\subsection{Subspace Learning}
\label{sec:4.1}

\subsubsection{Baselines}

In this section, we report the performance of L2-Graph for robust feature extraction. The competing methods include Locality Preserving Projections (LPP)~\cite{He2005Lap}, Neighborhood Preserving Embedding (NPE)~\cite{He2005}, Eigenfaces~\cite{Turk1991},  L1-Graph~\cite{Cheng2010}, Low Rank Representation (LRR)~\cite{Liu2013}, and Latent Low Rank Representation (LatLRR)~\cite{Liu2011}. We implemented a fast version of L1-Graph using Homotopy algorithm~\cite{Osborne2000} to compute the sparse representation. According to~\cite{Yang2010}, Homotopy is one of the most competitive $\ell_1$-minimization algorithms in terms of accuracy, robustness, and convergence speed. LRR and LatLRR are incorporated into the framework of NPE to obtain low-dimensional features similar to  L2-Graph and L1-Graph. After the low-dimensional features are extracted, we perform the nearest neighbor classifier to verify the performance of the tested methods.
In each test, we tuned the parameters of LPP, NPE, L1-Graph, LRR and LatLRR to achieve their best results. For L2-Graph, we fixed $\lambda=0.1$ and assigned different $k$ for different data sets. The used data sets and the MATLAB codes of L2-graph can be downloaded at \textcolor{blue}{\url{http://goo.gl/uAuISU}}.

\subsubsection{Data Sets}

Several popular facial data sets are used in our experiments, including Extended Yale Database B (ExYaleB)~\cite{Georghiades2001}, AR~\cite{Martinez1998}, and Multiple PIE (MPIE)~\cite{Gross2010}.

ExYaleB contains 2414 frontal-face images of 38 subjects (about 64 images for each subject), and we use the first 58 samples of each subject. The used AR data set contains 2600 samples from 50 male and 50 female subjects, of which 1400 samples are clean images, 600 samples are disguised by sunglasses, and the remaining 600 samples are disguised by scarves. MPIE contains the facial images captured in four sessions. In the experiments, all the frontal faces with 14 illuminations are investigated. For computational efficiency, we cropped each images from the original size to smaller one (see Table~\ref{tab:2}).

Each data set is partitioned into two parts, i.e., training data and testing data. Training data is used to learning a projection matrix, and the test datum is assigned to the nearest training datum in the projection feature space. For each algorithm, the same training and testing data partitions are used.

\begin{table}[t]
\caption{The used databases. $c$ and $n_i$ denote the number of subjects and the number of samples for each subject. }
\label{tab:2}
\begin{center}
\begin{small}
\begin{tabular}{lrllc}
\toprule
Databases &  $c$ & $n_i$ & Original Size & Cropped Size\\
\midrule
ExYaleB       &   38 & 58 & $192\times 168$ &  $54\times 48$ \\
AR1                &    100 & 26 & $165\times 120$ &  $55\times 40$ \\
AR2  & 100 & 12 & $165\times 120$ & $55\times 40$\\
AR3 & 100 & 12 & $165\times 120$ & $55\times 40$\\
MPIE-S1      &    249 & 14 & $100\times 82$ &  $55\times 40$  \\
MPIE-S2      &    203 & 10 & $100\times 82$ &  $55\times 40$  \\
MPIE-S3     &     164 & 10 & $100\times 82$ &  $55\times 40$  \\
MPIE-S4     &     176 & 10 & $100\times 82$ &  $55\times 40$  \\
COIL100 & 100 & 10 & $128\times128$& $64\times64$\\
\bottomrule
\end{tabular}
\end{small}
\end{center}
\end{table}

\subsubsection{Performance with Varying Training Sample and Feature Dimension}

In this section, we report the recognition results of L2-Graph over AR1 with increasing training data and ExYaleB with varying feature dimension. For the first test, we randomly selected $n_i$ AR images from each subject for training and used the rest for testing. Hence, we have $n_i$ training samples and $14 - n_i$ testing samples for each subject. For the second test, we split ExYaleB into two parts with equal size and perform 1-NN classifier over the first $m^{\prime}$ features, where $m^{\prime}$ increases from $1$ to $600$ with an interval of 10.

From~\figurename~\ref{fig:3}, one can conclude that: (1) L2-Graph performs well even though only a few of training data are available. Its accuracy is about $90\%$ when $n_i=4$, and the second best method achieve the same accuracy when $n_i=8$. (2) L2-Graph performs better than the other tested methods when $m^{\prime} \geq 50$. When more features are used ($m^{\prime} \geq350$), LRR and LatLRR are comparable to NPE and Eigenfaces which achieved the second and the third best result.

\subsubsection{Subspace Learning on Clean Facial Images}

In this section, the experiments are conducted  using MPIE. For each session of MPIE, we split it into two parts with the same data size. For each test, we fix $\lambda=0.1$ and $k=6$ for L2-Graph and tuned the parameters for the other algorithms.

\begin{table*}[t]
\caption{The recognition rate of 1-NN classifier with different subspace learning algorithms on the \textbf{MPIE database}. The values in parentheses denote the dimensionality of the features and the tuned parameters for the best result. The bold number indicates the best algorithm. }
\label{tab:3}
\begin{center}
\begin{small}
\begin{tabular}{crclrrrl}
\toprule
Databases          &	\multicolumn{1}{c}{L2-Graph}	      &		\multicolumn{1}{c}{Eigenfaces~\cite{Turk1991}} &	\multicolumn{1}{c}{LPP~\cite{He2005Lap}}	          &	\multicolumn{1}{c}{NPE~\cite{He2005}}	          &		\multicolumn{1}{c}{L1-Graph~\cite{Cheng2010}}	              &		\multicolumn{1}{c}{LRR~\cite{Liu2013}}		       &	\multicolumn{1}{c}{LatLRR~\cite{Liu2011}}\\
\midrule
MPIE-S1	           & \textbf{99.7}(249)  &	61.7(559)     &	53.4(595, 4) &	81.8(599,49) &	51.0(596,1e-3, 0.3) &	97.2(588,0.9) &	95.9(529,0.10)\\
MPIE-S2	           & \textbf{100.0}(243) &	47.9(272)     &	61.9(478, 2) & 92.8(494,49) &	94.1(544,1e-2, 0.1) &	99.8(380,1.0) &	99.3(486,0.10)\\
MPIE-S3	           & \textbf{99.9}(170)  &	42.8(556)     &	57.9(327,75) &	89.5(403,45) &	87.3(573,1e-3, 0.1) &	99.3(434,0.9) &	98.7(435,0.01)\\
MPIE-S4	           & \textbf{100.0}(175)   &	45.2(215)     &	60.3(398, 3) & 93.4(438,43) &	92.3(574,1e-3, 0.1) &	99.7(374,1.0) &	99.2(288,0.10)\\
\bottomrule
\end{tabular}
\end{small}
\end{center}
\end{table*}

\tablename~\ref{tab:3} reports the results. One can find that L2-Graph outperforms the other investigated approaches. The proposed method achieved $100\%$ recognition rates on the second and the third sessions of MPIE. In fact, it could have also achieved perfect classification results on MPIE-S1 and MPIE-S4 if different $\lambda$ and $k$ are allowed. Moreover, L2-Graph uses less dimensions but provides more discriminative  information.

\subsubsection{Subspace Learning on Corrupted Facial Images}

\begin{figure}
\begin{center}
\includegraphics[width=0.36\textwidth]{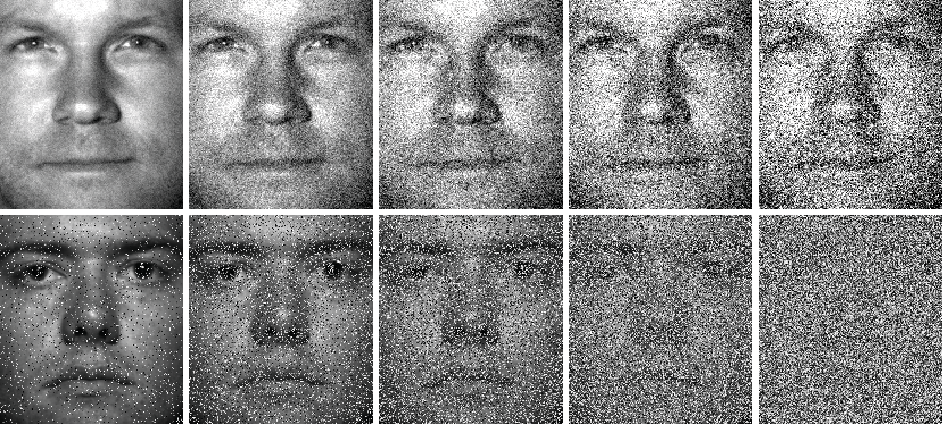}
\end{center}
\caption{\label{fig:4} The samples with real possible corruptions. Top row: the images with white Gaussian noise; Bottom row: the images with random pixel corruption. From left to right, the corruption rate increases from $10\%$ to $90\%$ (with an interval of $20\%$).}
\end{figure}

In this section, we investigate the robustness of L2-Graph ($\lambda=0.1$ and $k=15$) against two popular corruptions using ExYaleB over 38 subjects, i.e., white Gaussian noise (additive noise) and random pixel corruption (non-additive noise).~\figurename~\ref{fig:4} illustrates some samples.

\begin{table*}[t]
\caption{The recognition rate of the tested algorithms on the \textbf{Corrupted ExYaleB database}. WGN and RPC are shorts for White Gaussian Noise and Random Pixel Corruption, respectively. }
\label{tab:4}
\begin{center}
\begin{small}
\begin{tabular}{llclllll}
\toprule
Databases          &	\multicolumn{1}{c}{L2-Graph}	      &		\multicolumn{1}{c}{Eigenfaces~\cite{Turk1991}} &	\multicolumn{1}{c}{LPP~\cite{He2005Lap}}	          &	\multicolumn{1}{c}{NPE~\cite{He2005}}	          &		\multicolumn{1}{c}{L1-Graph~\cite{Cheng2010}}	              &		\multicolumn{1}{c}{LRR~\cite{Liu2013}}		       &	\multicolumn{1}{c}{LatLRR~\cite{Liu2011}}\\
\midrule
WGN+10\%	    & \textbf{95.2}(447)	& 79.4(474) &	82.7(495,2) & 94.0(527,49) &	84.9(558,\hspace{1.5mm}0.1,0.1) & 92.0(385,0.7) & 91.1(384,0.01)\\
WGN+30\%    & \textbf{92.1}(305)	& 70.5(128) &	71.9(444,2) & 87.9(343,47) &	72.3(451,1e-3,0.1) & 87.4(370,0.5) & 85.2(421,0.01)\\
RPC+10\%	& \textbf{87.0}(344)	& 69.8(\hspace{1mm}96) &	57.5(451,3) & 81.2(348,49) &	59.4(440,1e-3,0.1) & 80.5(351,0.5) & 77.1(381,0.10)\\
RPC+30\%	& \textbf{68.5}(332)	& 61.1(600) &	45.8(378,2) & 61.6(481,49) &	48.6(449,1e-3,0.1) & 58.9(361,0.5) & 57.2(364,0.01)\\
\bottomrule
\end{tabular}
\end{small}
\end{center}
\end{table*}

In the tests, we randomly chose half of the images (29 images per subject) to add these two types of corruptions.  Specifically, we added white Gaussian noise to the chosen sample $\mathbf{x}$ via $\mathbf{\tilde{x}} = \mathbf{x}+\rho \mathbf{n}$, where $\mathbf{\tilde{x}}\in[0\ 255]$, $\rho$ is the corruption ratio, and $\mathbf{n}$ is the noise following the standard normal distribution. For non-additive corruption, we replace the value of a percentage of pixels randomly selected from the image with the values following an uniform distribution over $[0,\ p_{max}]$, where $p_{max}$ is the largest pixel value of $\mathbf{x}$.

From \tablename~\ref{tab:4}, it is easy to find that L2-Graph is superior to the other approaches with  a considerable performance gain. When $30\%$ pixels are randomly corrupted, the accuracy of L2-Graph is at least $6.9\%$ higher than that of the other methods.

\subsubsection{Subspace Learning on Disguised Facial Images}

\begin{figure}
\begin{center}
\includegraphics[width=0.45\textwidth]{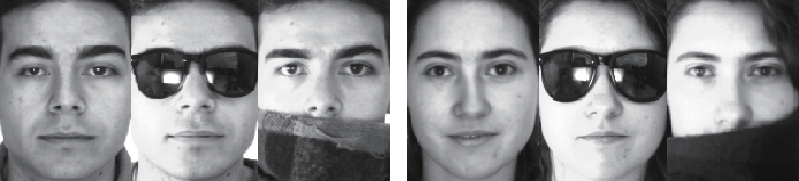}
\end{center}
\caption{\label{fig:5} Some sample images disguised by sunglasses (AR2) and scarves (AR3).}
\end{figure}

\tablename~\ref{tab:5} reports the results of L2-Graph ($\lambda=0.1$ and $k=3$) over two subsets of AR database (\figurename~\ref{fig:5}). The first subset (AR2) contains 600 clean images and 600 disguised images with sunglasses (occlusion rate is about $20\%$), and the second one (AR3) includes 600 clean images and 600 images disguised by scarves (occlusion rate is about $40\%$). L2-Graph again outperforms the other tested methods by a considerable performance margin. With respect to two different disguises, the recognition rates of L2-Graph are $5.8\%$ and $7.5\%$ higher than those of the second best method.

\begin{table}[t]
\caption{Classification performance of the tested algorithms on the \textbf{disguised AR images}.}
\label{tab:5}
\begin{center}
\begin{small}
\begin{tabular}{lll}
\toprule
Algorithms &AR2 (sunglasses) & AR3 (scarves)\\
\midrule
L2-Graph & \textbf{85.3}(479) & \textbf{83.3}(585)\\
Eigenfaces~\cite{Turk1991} 	& 35.7(494) & 33.5(238)\\
LPP~\cite{He2005Lap}	            & 44.2(228,85) & 40.7(222,95)\\
NPE~\cite{He2005}            & 54.9(120,45) & 61.2(284,49)\\
L1-Graph~\cite{Cheng2010}            & 78.5(598,1e-2,0.1) & 72.0(589,1e-3,0.1)\\
LRR~\cite{Liu2013}            & 79.2(590,1e-7) & 75.8(591,1.0)\\	
LatLRR~\cite{Liu2011}      &	79.5(593,0.1) & 74.0(600,1e-5)\\
\bottomrule
\end{tabular}
\end{small}
\end{center}
\end{table}

\subsection{Image Clustering}
\label{sec:4.2}

\begin{figure*}[t]
\begin{center}
\subfigure []{\label{fig:6a}\includegraphics[width=0.46\textwidth]{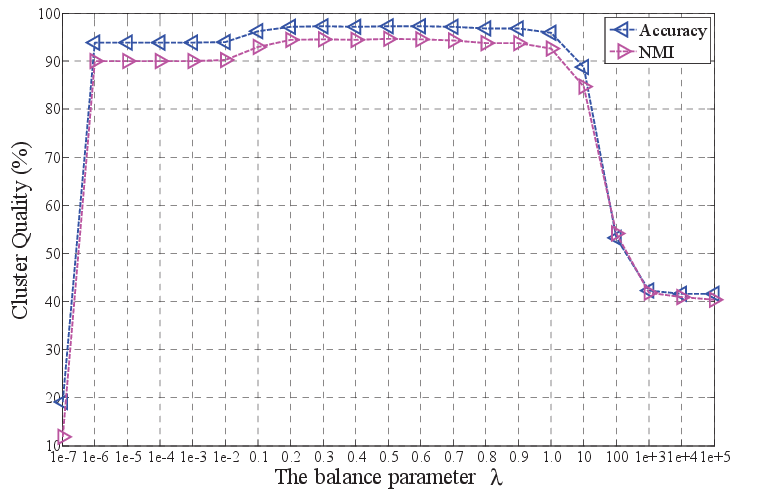}}
\subfigure []{\label{fig:6b}\includegraphics[width=0.46\textwidth]{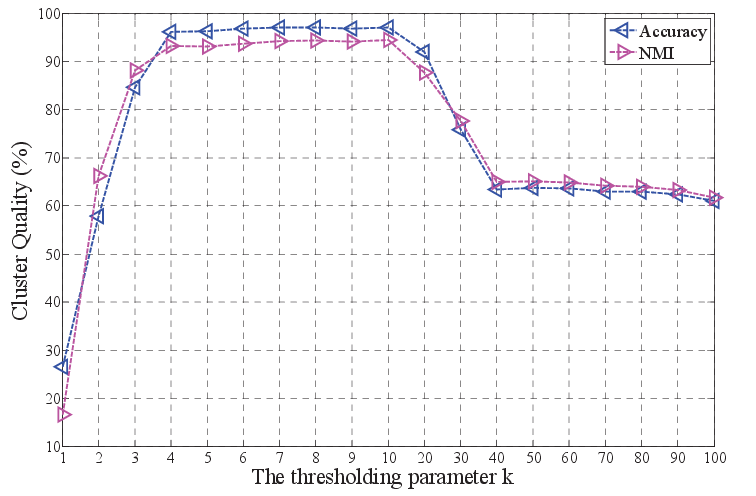}}
\end{center}
\caption{\label{fig:6} The influence the parameters of L2-Graph. (a) The influence of $\lambda$, where $k=7$. (b) The influence of $k$, where $\lambda=0.7$. One can find that, L2-Graph successfully eliminates the effect of errors by keeping $k$ largest entries. The example verifies the effectiveness of our theoretical results.}
\end{figure*}

\subsubsection{Baselines} We compared L2-Graph with several recently-proposed subspace clustering algorithms, i.e., SSC~\cite{Elhamifar2013}, LRR~\cite{Liu2013}, and two variants of LSR (LSR1 and LSR2)~\cite{Lu2012}. Moreover, we used the coefficients of Locally Linear Embedding (LLE)~\cite{Roweis2000} to build the similarity graph for subspace clustering as~\cite{Cheng2010} did, denoted by LLR (i.e., Locally Linear Representation).

For fair comparison, we performed the same spectral clustering algorithm~\cite{Ng2002} on the graphs built by the tested algorithms and reported their best results with the tuned parameters. For the SSC algorithm, we experimentally found an optimal $\alpha$ from $1$ to $50$ with an interval of 1. For LRR, the optimal $\lambda$ was found from $10^{-6}$ to 10 as suggested in~\cite{Liu2013}. For LSR1, LSR2, and L2-Graph, the optimal $\lambda$ was chosen from $10^{-7}$ to 1. Moreover, a good $k$ was found from 3 to 14 for L2-Graph and from 1 to 100 for LLR.

\subsubsection{Evaluation Metrics} Two popular benchmarks, \emph{Accuracy} (or called \emph{Purity}) and Normalized Mutual Information (\emph{NMI})~\cite{Cai2005}, are used to evaluate the clustering quality. The value of \emph{Accuracy} or \emph{NMI} is 1 indicates perfect matching with the ground truth, whereas 0 indicates perfect mismatch.

\subsubsection{Data Sets}

We investigate the performance of the methods on the data sets summarized in \tablename~\ref{tab:2}. For computational efficiency, we downsized each image from the original size to a smaller one and performed Principle Component Analysis (PCA) to reduce the dimensionality of the data by reserving $98\%$ energy. For example, all the AR1 images were downsized and normalized from $165\times 120$ to $55\times 40$. After than, the experiment were carried out using 167 features extracted by PCA.

\subsubsection{Model Selection}

 L2-Graph has two parameters, the balance parameter $\lambda$ and the thresholding parameter $k$. The values of these parameters depend on the data distribution. In general, a bigger $\lambda$ is more suitable to characterize the corrupted images and $k$ equals to the dimensionality of the corresponding subspace.

To examine the influence of these parameters, we carried out some experiments using a subset of ExYaleB which contains 580 images from the first 10 individuals. We randomly selected a half of samples to corrupt using white Gaussian noise.~\figurename~\ref{fig:6} shows that:
\begin{itemize}
  \item while $\lambda$ increases from $0.1$ to $1.0$ and $k$ ranges from $4$ to $9$, \emph{Accuracy} and \emph{NMI} almost remain unchanged;
  \item the thresholding parameter $k$ is helpful to improve the robustness of our model. This verifies the correctness of our theoretical result that the trivial coefficients correspond to the codes over the errors, i.e., IPD property of $\ell_2$-norm based projection space;
  \item a larger $k$ will impair the discrimination of the model, whereas a smaller $k$ cannot provide enough representative ability. Indeed, the optimal value of $k$ can be found around the intrinsic dimensionality of the corresponding subspace. According to~\cite{Costa2004}, the intrinsic dimensionality of the first subject of Extended Yale B is 6. This result is consistent with our experimental result.
\end{itemize}

\subsubsection{Performance with Varying Number of Subspace}

\begin{figure*}[t]
\begin{center}
\subfigure [\emph{Accuracy}]{\label{fig:7a}\includegraphics[width=0.46\textwidth]{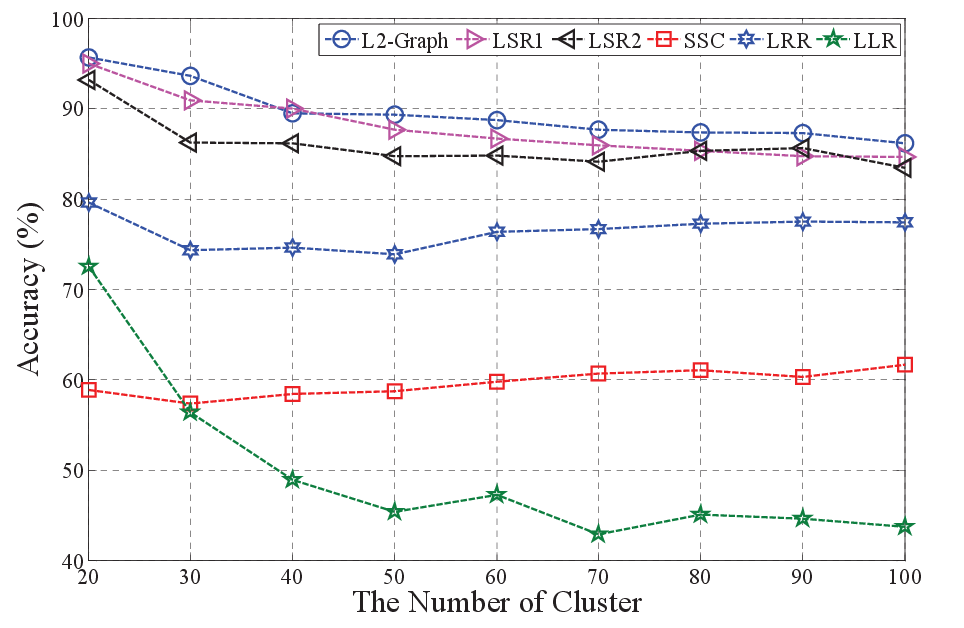}}
\subfigure [\emph{NMI}]{\label{fig:7b}\includegraphics[width=0.46\textwidth]{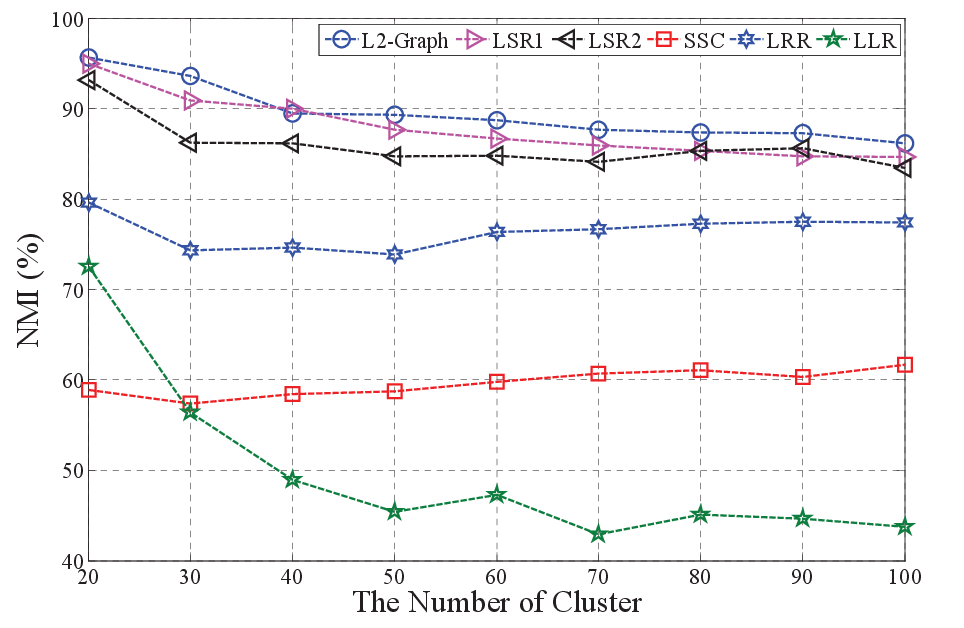}}
\end{center}
\caption{\label{fig:7} The clustering quality (\emph{Accuracy} and \emph{NMI}) of different algorithms on the first $c$ subjects of AR data set.}
\end{figure*}

 In this section, we evaluate the performance of L2-Graph using 1400 clean AR images (167 dimension). The experiments were carried out on the first $c$ subjects of the data set, where $c$ increases from $20$ to $100$. \figurename~\ref{fig:7} shows that:
 \begin{itemize}
  \item L2-Graph algorithm is more competitive than the other examined algorithms. For example, when $L=100$, the \emph{Accuracy} of L2-Graph is at least, $1.8\%$ higher than that of LSR1, $2.7\%$ higher than that of LSR2, $24.5\%$ higher than that of SSC , $8.8\%$ higher than that of LRR and $42.5\%$ higher than that of LLR;
  \item with increasing $c$, the $NMI$ of L2-Graph almost remain unchanged, slightly varying from $93.0\%$ to $94.3\%$. The possible reason is that $NMI$ is robust to the data distribution (increasing subject number).
\end{itemize}

\subsubsection{Clustering on Clean Images}

Six image data sets (ExYaleB, MPIE-S1, MPIE2-S2, MPIE3-S3, MPIE-S4, and COIL100) are used in this experiment. \tablename~\ref{tab:6} shows that
\begin{itemize}
  \item the L2-graph algorithm achieves the best results in the tests except with MPIE-S4, where it is second best. With respect to the ExYaleB database, the \emph{Accuracy} of the L2-graph is about 10.28\% higher than that of the LSR, 12.19\% higher than that of the LSR2, 18.18\% higher than that of the SSC, 1.53\% higher than that of the LRR, and 34.96\% higher than that of LLR;
  \item in the tests, L2-Graph, LSR1, and LSR2 exhibit similar performance, because the methods are $\ell_2$-norm-based methods. One of the advantages of L2-Graph is that it is more robust than LSR1, LSR2, and the other tested methods.
\end{itemize}

\begin{table*}[th]
\caption{Clustering Performance (\%)) on Six Different Image Data Sets. }
\label{tab:6}
\begin{center}
\begin{tabular}{c| ll | ll |  ll |ll | ll | ll}
\toprule
\multicolumn{1}{c|}{\multirow{2}{*}{Databases}} & \multicolumn{2}{c|}{L2-graph} & \multicolumn{2}{c|}{LSR1~\cite{Lu2012}} & \multicolumn{2}{c|}{LSR2~\cite{Lu2012}} & \multicolumn{2}{c|}{SSC~\cite{Elhamifar2013}} & \multicolumn{2}{c|}{LRR~\cite{Liu2013}} & \multicolumn{2}{c}{LLR~\cite{Roweis2000}} \\
\cline{2-13}
 & Accuracy & NMI & Accuracy & NMI & Accuracy & NMI & Accuracy & NMI & Accuracy & NMI & Accuracy & NMI \\
\midrule
ExYaleB &	\textbf{86.78}	(1.0,5)& \textbf{92.84}	& 76.50	(1e-3)	& 80.59	& 74.59	(1e-4)	& 79.05	& 68.60	(8)& 75.04	& 85.25	(10	) & 91.19	& 51.82	(3) & 61.61\\
MPIE-S1 &	\textbf{88.12}	(1e-3,7)& \textbf{96.75}	& 87.55	(0.01)	& 95.64	& 85.60	(1e-4)	& 95.35	& 68.39	(11) & 89.60	& 83.88	(0.7)	& 95.76	& 40.22	(100)	& 76.57\\
MPIE-S2 &	\textbf{90.76}	(1e-4,5)& \textbf{98.57}	& 89.79	(1e-4)	& 97.65	& 88.35	(1e-4)	& 96.52	& 76.60	(9) & 95.27	& 81.03	(5)	& 96.73	& 31.77	(60) & 74.21\\
MPIE-S3 &	\textbf{88.23}	(0.01,7)& 97.72	& 87.89	(0.01)	& 95.24	& 88.10	(0.01)	& \textbf{98.14}	& 66.83	(8) & 92.05	& 75.61	(0.7)	& 95.40	& 28.48	(5) & 72.44\\
MPIE-S4 &	90.51	(0.01,5)& 98.54	& 89.85	(0.01)	& 97.66	& \textbf{91.01}	(0.01)	& \textbf{98.91}	& 77.84	(13) & 95.31	& 83.24	(0.7)	& 97.09	& 42.96	(95) & 80.60\\
COIL100 &	\textbf{52.40}	(10,7)& \textbf{77.57}	& 50.70	(0.50)	& 76.05	& 49.60	(0.20)	& 75.94	& 51.40	(20) & 76.93	& 50.10	(0.1)	& 76.29	& 48.60	(8) & 75.30\\
\bottomrule
\end{tabular}
\end{center}
\end{table*}

\subsubsection{Clustering on Corrupted Images}

\begin{table*}[t]
\caption{The performance of L2-Graph, LSR~\cite{Lu2012}, SSC~\cite{Elhamifar2013}, LRR~\cite{Liu2013}, and LLR~\cite{Roweis2000} on the \textbf{ExYaleB} database (116 dimension). $\rho$ denotes the corrupted ratio; The values in the parentheses denote the optimal parameters for the reported \emph{Accuracy}, i.e., L2-Graph ($\lambda$, $k$), LSR ($\lambda$), SSC($\alpha$), LRR ($\lambda$), and LLR ($k$).}
\label{tab:7}
\begin{center}
\begin{footnotesize}
\begin{tabular}{c|c|ll|ll|ll|ll|ll|ll}
\toprule
\multicolumn{1}{c|}{\multirow{2}{*}{Corruption}} & \multicolumn{1}{c|}{\multirow{2}{*}{$\rho$}} & \multicolumn{2}{c|}{L2-Graph} & \multicolumn{2}{c|}{LSR1~\cite{Lu2012}} & \multicolumn{2}{c|}{LSR2~\cite{Lu2012}} &  \multicolumn{2}{c|}{SSC~\cite{Elhamifar2013}} & \multicolumn{2}{c|}{LRR~\cite{Liu2013}} & \multicolumn{2}{c}{LLR~\cite{Roweis2000}}\\
\cline{3-14}
 &  & Accuracy & NMI & Accuracy & NMI & Accuracy & NMI & Accuracy & NMI & Accuracy & NMI & Accuracy & NMI\\
\midrule
 & 10 & \textbf{89.25}(1e-4,6) & \bfseries92.71 & 72.28(1e-2) & 78.36 & 73.19(1e-4) & 78.52 &
68.38(8) & 74.25 & 87.79(0.7) & 92.12 & 47.82(5) & 69.40  \\
White & 30 & \textbf{88.70}(0.7,6) & \bfseries92.18 & 71.14(1e-4) & 75.93 & 74.55(1e-4) & 78.30 &
66.02(10) & 71.50 & 81.31(5.0) & 86.05 & 46.51(6) & 59.84  \\
Gaussian & 50 & \textbf{86.57}(0.7,4) & \bfseries90.43 & 63.61(1e-2) & 70.58 & 63.16(1e-4) & 71.79 &
55.85(22) & 61.99 & 84.96(0.4) & 79.15 & 37.48(5) & 52.10    \\
Noise & 70 & \textbf{74.32}(0.6,7) & \bfseries77.70 & 52.72(1e-3) & 63.08 & 51.54(1e-4) & 63.02 &
49.00(30) & 58.64 & 60.66(0.7) & 69.57 & 32.76(5) & 44.96    \\
 & 90 & \textbf{56.31}(0.6,7) & \bfseries63.43 & 43.15(0.1) & 55.73 & 42.33(1e-4) & 55.64 &
44.10(36) & 51.79 & 49.96(0.2) & 57.90 & 29.81(5) & 42.90    \\
\hline
& 10 & \textbf{82.76}(1.0,4) & \bfseries88.64 & 72.35(1e-3) & 77.09 & 72.35(1e-4) & 77.11 &
64.97(48) & 68.40 & 78.68(0.3) & 87.19 & 46.82(6) & 59.26  \\
 Random & 30 & \textbf{68.97}(0.7,7) & \bfseries75.89 & 56.48(1e-4) & 63.19 & 56.48(1e-2) & 63.28 &
56.13(49) & 59.96 & 60.80(0.6) & 67.47 & 33.26(5) & 42.33    \\
 Pixels & 50 & \textbf{48.15}(1.0,6) & \bfseries56.67  & 42.15(1e-4) & 50.53 & 43.16(0.4) & 53.09 &
45.60(39) & 51.69 & 38.61(0.2) & 49.93 & 19.51(5) & 27.77   \\
 Corruption & 70 & \textbf{34.98}(1e-2,5) & \bfseries45.56 & 27.86(1e-3) & 35.88 & 27.50(1e-2) & 35.73 &
34.71(48) & 41.14 & 30.54(0.2) & 38.13 & 13.39(6) & 18.82   \\
 & 90 & \textbf{30.04}(1e-4,4) & \bfseries38.39   & 19.78(1e-3) & 28.00 & 19.19(0.1) & 28.22 &
20.78(47) & 30.03 & 19.01(0.2) & 29.16 & 14.07(6) & 23.04 \\
\bottomrule
\end{tabular}
\end{footnotesize}
\end{center}
\end{table*}

 Our error removing strategy can improve the robustness of L2-Graph without the prior knowledge of the errors. To verify this claim, we test the robustness of L2-Graph using ExYaleB over $38$ subjects. For each subject of the database, we randomly chose a half of images (29 images per subject) to corrupt by white Gaussian noise or random pixel corruption, where the former is additive and the latter is non-additive. To avoid randomness, we produced ten data sets beforehand and then performed the evaluated algorithms over these data partitions. From \tablename~\ref{tab:7}, we have the following conclusions:
\begin{itemize}
  \item all the investigated methods perform better in the case of white Gaussian noise. The result is consistent with a widely-accepted conclusion that non-additive corruptions are more challenging than additive ones in pattern recognition;
  \item L2-Graph is again considerably more robust than LSR1, LSR2, SSC, LRR and LLR. For example, with respect to white Gaussian noise, the performance gain in \emph{Accuracy} between L2-Graph and LSR2 varied from $14.0\%$ to $22.8\%$; with respect to random pixel corruption, the performance gain varied from $5.0\%$ to $13.2\%$.
\end{itemize}

\begin{figure*}[t]
\centering
\includegraphics[width=0.98\textwidth]{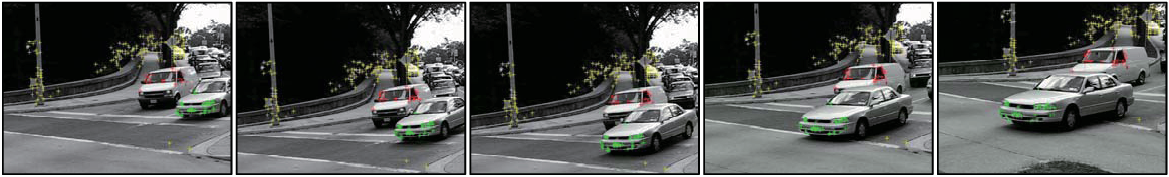}
\caption{\label{fig:8} Some sample frames taken from the Hopkins155 database.}
\end{figure*}

\begin{table*}[t]
\caption{Clustering performance of different methods on the \textbf{disguised AR images}. The values in parentheses denote the optimal parameters for accuracy.}
\label{tab:8}
\begin{center}
\begin{small}
\begin{tabular}{l |lll|lll}
\toprule
\multicolumn{1}{c|}{\multirow{2}{*}{Algorithms}} & \multicolumn{3}{c|}{Occluded by sunglasses} & \multicolumn{3}{c}{Occluded by scarves} \\
\cline{2-7}
& \multicolumn{1}{c}{\multirow{1}{*}{Accuracy}} & \multicolumn{1}{c}{\multirow{1}{*}{NMI}} & \multicolumn{1}{c|}{\multirow{1}{*}{Time (s)}} & \multicolumn{1}{c}{\multirow{1}{*}{Accuracy}} & \multicolumn{1}{c}{\multirow{1}{*}{NMI}} & \multicolumn{1}{c}{\multirow{1}{*}{Time (s)}}\\
\midrule
L2-Graph  & \textbf{75.92}  & \bfseries 88.73 & \bfseries 73.63 & \textbf{79.08}  & \bfseries 89.61 & \bfseries 89.53\\
LSR1~\cite{Lu2012} & 72.83 (1e-4) & 84.48 & 126.85 & 75.75 (1e-3) & 88.53 & 132.65\\
LSR2~\cite{Lu2012} & 73.75 (1e-3) & 86.81 & 128.45 & 74.67 (1e-3) & 87.91 & 132.72\\
SSC-Homotopy    & 45.33 (1e-7,1e-3)         & 73.81           & 306.99           & 38.83 (1e-7,1e-3)        & 70.84           & 353.92\\
SSC~\cite{Elhamifar2013}         & 35.75 (36)                & 67.64           & 376.23           & 35.08 (48)               & 68.30           & 276.07 \\
LRR~\cite{Liu2013}       & 62.00 (5)                 & 84.81           & 226.93           & 61.50 (10)               & 82.88           & 215.76\\
LLR~\cite{Roweis2000}       & 27.33 (95)                & 61.28           & 296.88           & 25.67 (85)               & 59.15           & 304.66\\
\bottomrule
\end{tabular}
\end{small}
\end{center}
\end{table*}

\subsubsection{Clustering on Disguised images}

In this section, we examine the robustness to real possible occlusions of the competing methods by using AR2 and AR3. Beside the implementation of Elhamifar et\ al.~\cite{Elhamifar2013}, we also report the result by using Homotopy method~\cite{Osborne2000} to solve the $\ell_1$-minimization problem. In the experiments, we fix $\lambda=0.001$ and $k=12$ for the L2-Graph and tuned the parameters of the tested methods for achieving their best performance.

\tablename~\ref{tab:3} reports the performance of the tested algorithms. Clearly, L2-Graph again outperforms the other methods in clustering quality and efficiency. Its $Accuracy$ is about $30.59\%$ higher than SSC-Homotopy, $40.17\%$ higher than SSC, $13.92\%$ higher than LRR and $48.59\%$ higher than LLR when the faces are occluded by glasses. In the case of the faces occluded by scarves, the figures are about $40.25\%$, $44.00\%$, $17.58\%$ and $54.31\%$, respectively. In addition, we can see that each of the evaluated algorithm performs very close for two different disguises, even though the occluded rates are largely different.

\subsection{Motion Segmentation}

Motion segmentation aims to separate a video sequence into multiple spatiotemporal regions of which each region represents a moving object. Generally, segmentation algorithms are based on the feature point trajectories of multiple moving objects. Therefore, the motion segmentation problem can be thought of the clustering of these trajectories into different subspaces, and each subspace corresponds to an object.

To examine the performance of the proposed approach for motion segmentation, we conducted experiments on the Hopkins155 raw data~\cite{Tron2007}, some frames of which are shown in \figurename~\ref{fig:8}. The data set includes the feature point trajectories of 155 video sequences, consisting of 120 video sequences with two motions and 35 video sequences with three motions. Thus, there are a total of 155 independent clustering tasks. For each algorithm, we report the mean and median of clustering errors (1-$Accuracy$) using these two data partitions (two and three motions). For L2-Graph, we fixed  $\lambda=0.1$ and $k=7$ ($k=14$) for AR2 and AR3. For the other methods, we tuned the parameters by following experimental setups in~\cite{Elhamifar2013,Liu2013,Lu2012}.

\begin{table}[t]
\caption{Segmentation errors (\%) on the Hopkins155 raw data.}
\label{tab:9}
\begin{center}
\begin{small}
\begin{tabular}{l|lr|lr}
\toprule
\multicolumn{1}{c|}{\multirow{2}{*}{Methods}} & \multicolumn{2}{c|}{\multirow{1}{*}{2 motions}} & \multicolumn{2}{c}{\multirow{1}{*}{3 motions}}\\
\cline{2-5}
 &  \multicolumn{1}{c}{mean}  &  \multicolumn{1}{c|}{median}  &  \multicolumn{1}{c}{mean}  &  \multicolumn{1}{c}{median} \\
\midrule
L2-Graph       &\hspace{1mm}2.53  & \textbf{0.00} & \hspace{1mm}\textbf{6.16}  & \textbf{1.00} \\
LSR1~\cite{Lu2012}      & \hspace{1mm}3.16 (4.6e-3) & 0.27 &  \hspace{1mm}6.50 (4.6e-3) & 2.05 \\
LSR2~\cite{Lu2012}      & \hspace{1mm}3.13 (4.8e-3) & 0.22 & \hspace{1mm}6.94 (4.6e-3) & 2.05 \\
SSC~\cite{Elhamifar2013}       & \hspace{1mm}4.63 (1e-3) & 0.61 &  \hspace{1mm}8.77 (1e-3) & 5.29 \\
LRR~\cite{Liu2013}       & \hspace{1mm}\textbf{2.22} (0.4) & 0.00 &  \hspace{1mm}7.45 (0.7) & 1.57 \\
LLR~\cite{Roweis2000} & 12.46 (9) & 3.28 &  19.62 (6) & 18.95 \\
\bottomrule
\end{tabular}
\end{small}
\end{center}
\end{table}

\tablename~\ref{tab:9} reports the mean and median segmentation errors on the data sets. We can find that the L2-Graph outperforms the other tested methods on the three-motions data set and performs comparable to the methods on two-motion case. Moreover, all the algorithms perform better with two-motion data than with three-motion data.

\section{Conclusion}
\label{sec:6}

Under the framework of graph-based learning, most of the recent approaches achieve robust clustering results by removing the errors from the original space and then build the neighboring relation based on a `clean' data set. In contrast, we propose and prove that it is feasible to eliminate the effect of the errors from the linear projection space (representation). Based on this mathematically traceable property (called Intra-subspace Projection Dominance), we present two simple but effective methods for robust subspace learning and clustering. Extensive experimental results validate the excellent performance of our approach in unsupervised feature extraction, image clustering, and motion segmentation.

There are several ways to further improve or extend this work. Although the theoretical analysis and experimental studies showed connections between the parameter $k$ and the intrinsic dimensionality of a subspace, it is challenging to determine the optimal value of the parameter. Therefore, we intend to explore more theoretical results on model selection in the future.

\ifCLASSOPTIONcaptionsoff
  \newpage
\fi

\bibliographystyle{IEEEtran}
\bibliography{IEEEabrv,L2graphBib}

\end{document}